\documentclass{IEEEtaes}

\usepackage{cite}
\usepackage{amsmath,amssymb,amsfonts}
\usepackage{graphicx}
\usepackage{textcomp}
\usepackage{xcolor}
\usepackage{algorithm}
\usepackage{algorithmic}
\usepackage{bm}
\usepackage{comment}
\usepackage{booktabs}
\usepackage{multirow}
\usepackage{siunitx}
\newcommand{\nrf}{D}
 \newcommand{\bbbeta}{\bm{\beta}}
\newcommand{\bbtheta}{\bm{\theta}}
\newcommand{\bbSig}{\bm{\Sigma}}
\newcommand{\bbphi}{\boldsymbol{\phi}} 
\newcommand{\change}[1]{{\color{red}#1}}

\def\BibTeX{{\rm B\kern-.05em{\sc i\kern-.025em b}\kern-.08em
    T\kern-.1667em\lower.7ex\hbox{E}\kern-.125emX}}

\jvol{XX}
\jnum{XX}
\jmonth{XXXXX}
\paper{1234567}
\pubyear{2020}
\doiinfo{TAES.2020.Doi Number}

\begin{document}

\title{Adaptive Bayesian Optimization for Robust Identification of Stochastic Dynamical Systems }

\author{Jinwen Xu}
\member{Student Member, IEEE}
\affil{School of ECE, University of Georgia, Athens, GA, USA} 

\author{Qin Lu}
\member{Member, IEEE}
\affil{School of ECE, University of Georgia, Athens, GA, USA} 

\author{Yaakov Bar-Shalom}
\member{Life Fellow, IEEE}
\affil{Department of ECE, University of Connecticut, Storrs, CT, USA} 


\receiveddate{Manuscript received XXXXX 00, 0000; revised XXXXX 00, 0000; accepted XXXXX 00, 0000.\\
J. Xu and Q. Lu are supported by NSF \#2340049.}


\authoraddress{The next few paragraphs should contain the authors' current affiliations, including current address and e-mail. For example, First A. Author is with the National Institute of Standards and Technology, Boulder, CO 80305 USA 
(e-mail: \href{mailto:author@boulder.nist.gov}{author@boulder.nist.gov}). Second B. Author, Jr., was with Rice University, Houston, TX 77005 USA. He is now with the Department of Physics, Colorado State University, Fort Collins, CO 80523 USA (e-mail: \href{mailto:author@lamar.colostate.edu}{author@lamar.colostate.edu}). Third C. Author is with the Electrical Engineering Department, University of Colorado, Boulder, CO 80309 USA, on leave from the National Research Institute for Metals, Tsukuba 305-0047, Japan 
(e-mail: \href{mailto:author@nrim.go.jp}{author@nrim.go.jp}).}


\markboth{AUTHOR ET AL.}{SHORT ARTICLE TITLE}
\maketitle

\begin{abstract}
This paper deals with the identification of linear stochastic dynamical systems, where the unknowns include system coefficients and noise variances. Conventional approaches that rely on the maximum likelihood estimation (MLE) require nontrivial gradient computations and are prone to local optima.  To overcome these limitations, a sample-efficient global optimization method based on Bayesian optimization (BO) is proposed, using an ensemble Gaussian process (EGP) surrogate with weighted kernels from a predefined dictionary.  This ensemble enables a richer function space and improves robustness over single-kernel BO.  Each objective evaluation is efficiently performed via Kalman filter recursion.  Extensive experiments across parameter settings and sampling intervals show that the EGP-based BO consistently outperforms MLE via steady-state filtering and expectation-maximization (whose derivation is a side contribution) in terms of RMSE and statistical consistency.  Unlike the ensemble variant, single-kernel BO does not always yield such gains, underscoring the benefits of model averaging.  Notably, the BO-based estimator achieves RMSE below the classical Cramér–Rao bound, particularly for the inverse time constant, long considered difficult to estimate.  This counterintuitive outcome is attributed to a {\it data-driven prior} implicitly induced by the GP surrogate in BO.

\end{abstract}


\section{Introduction}
Stochastic dynamical systems, governed by stochastic differential equations (SDEs), play a central role in modeling time-evolving phenomena in fields such as signal processing, control theory, finance, and biology~\cite{ljung1994modeling,anderson2012optimal,BarShalom2001Estimation}. These models capture complex temporal dependencies, including memory effects, inertia, and higher-order dynamics, by incorporating derivatives beyond the first order~\cite{Oksendal2003SDE}. In continuous time, such systems are typically driven by Gaussian white noise and observed through noisy measurements~\cite{BarShalom2001Estimation}. Discretized versions are often used for inference and prediction when working with sampled data~\cite{anderson2012optimal, Oksendal2003SDE}.

A classical example is the Ornstein–Uhlenbeck (OU) process, which corresponds to the first-order case and has been extensively used to model mean-reverting behaviors. In biology and ecology, the OU process enhances Brownian motion by introducing stabilizing selection toward optimal trait values \cite{hansen1997stabilizing}. Its widespread application is supported by tools such as the ouch and GEIGER R packages \cite{butler2004phylogenetic,harmon2008geiger}, enabling inference in phylogenetic niche conservatism, convergent evolution, and adaptive radiation \cite{wiens2010niche,ingram2013surface}. More complex biological or physical systems may require second-order or higher-order SDEs to properly capture dynamic dependencies such as acceleration or feedback.

\noindent {\bf Related works.}
The performance of high-order stochastic dynamical systems critically depends on the accurate estimation of their parameters, including system coefficients and the variances of process and observation noise. Based on the log-likelihood function (LLF) as the objective, a maximum likelihood estimation (MLE) problem can be formulated. However, the LLF is typically highly nonlinear and nonconvex, making its gradient difficult to evaluate directly~\cite{nikolic2015maximum}. The expectation-maximization (EM) algorithm provides an alternative by avoiding explicit gradient computations, but its convergence is sensitive to initialization and only guarantees local optima~\cite{stebler2011constrained,wu1983convergence}. More recently, a steady-state Kalman filter (KF) approximation has been employed for efficient MLE in the first-order setting~\cite{ye2024maximum}, along with the derivation of the classical Cramér–Rao lower bound (CRLB) to benchmark estimation accuracy. Nonetheless, the estimation of certain parameters—such as inverse time constants or high-order coefficients—remains challenging, particularly under limited data or low signal-to-noise regimes~\cite{roy2001estimation,thomas2016bias}.

Recent developments in SDE parameter estimation include asymptotically efficient methods for hidden Ornstein-Uhlenbeck processes using Kalman-Bucy filtration~\cite{kutoyants2019parameter} and multi-step MLE processes for ergodic diffusions that achieve near-optimal performance with reduced computational cost~\cite{kutoyants2017multistep}. Alternative approaches in the SDE literature include method of moments~\cite{duffie1993simulated,gallant1996efficient} and Bayesian methods~\cite{golightly2011bayesian}, which have shown success in various applications. However, these established methods face significant challenges when applied to second-order and higher-order stochastic dynamical systems. The computational burden scales unfavorably with system dimensionality, as the state space includes multiple derivatives and their interactions. Moreover, the theoretical guarantees and convergence properties derived for first-order SDE models may not hold for the complex coupling structures inherent in second-order and higher-order dynamics. The fundamental choice between MLE and EM for likelihood-based estimation of second-order and higher-order systems remains understudied, particularly regarding their relative computational efficiency, estimation accuracy, and robustness across different system orders and noise conditions.

\noindent{\bf Contributions.} 
To enable consistent identification of high-order stochastic dynamical systems with convergence to global optimum, a novel Bayesian optimization (BO) based approach is advocated, where the negative log-likelihood (NLL) is treated as a black-box objective and approximated using a Ensemble Gaussian process (EGP) surrogate. The main contributions are summarized as follows:

\begin{enumerate}
    \item[c1)] Relying on the BO framework, a novel approach is developed for the identification of a general continuous-time linear stochastic systems, with state-space representations obtained via exact matrix exponential discretization. Both first-order (Ornstein–Uhlenbeck) and second-order models are considered as illustrative examples.

    \item[c2)] A weighted ensemble of GP surrogates, each with a distinct kernel from a predefined dictionary, is employed to approximate the NLL. This ensemble surrogate captures a richer function space than standard single-kernel approaches and improves robustness and accuracy across heterogeneous scenarios.

    \item[c3)] Extensive simulation studies across diverse parameter settings and sampling intervals demonstrate that the proposed BO-based estimator—when equipped with the EGP surrogate—consistently outperforms MLE (based on steady-state Kalman filtering) and the expectation-maximization (EM) algorithm (whose derivation is included as a side contribution), in terms of root mean-square error (RMSE), normalized estimation error squared (NEES), and normalized innovation squared (NIS). Particularly, RMSE for challenging parameters such as the inverse time constant is often found to fall below the classical Cramér–Rao lower bound (CRLB)~\cite{ye2024maximum}, which is explained by the {\it data-driven prior} implicitly introduced via the EGP surrogate, yielding a Bayesian CRLB that differs from the classical counterpart.
\end{enumerate}

\noindent The closed-form posterior mean and variance from the GP surrogates enable principled acquisition functions that guide efficient and globally optimal parameter search, offering a sample-efficient alternative to conventional likelihood-based methods.

\section{Problem Formulation}

We consider a scalar $n$-th order stochastic differential equation (SDE) of the form
\begin{align}
    x^{(n)}(t) + a_{n-1}x^{(n-1)}(t) + \cdots + a_1\dot{x}(t) + a_0x(t) = \tilde{v}(t), \label{eq:ho_sde}
\end{align}
where $x(t)$ is the latent state, and $\tilde{v}(t)$ is a zero-mean white process noise with autocorrelation
\begin{align}
    \mathbb{E}[\tilde{v}(t)\tilde{v}(\tau)] = \tilde{Q} \delta(t - \tau),
\end{align}
where $\tilde{Q}$ denotes the power spectral density (PSD) of the driving noise. Our goal is to develop an equivalent discrete-time state-space model suitable for parameter estimation.

To gain analytical insight into the system behavior, we begin by expressing the solution using the Green’s function associated with the differential operator~\cite{stakgold2011green,cole1968perturbation}. The general solution can be written as
\begin{align}
    x(t) = x_{\text{hom}}(t) + \int_0^t G(t - \tau) \tilde{v}(\tau) d\tau,
\end{align}
where $x_{\text{hom}}(t)$ is the homogeneous solution, and $G(t)$ is the impulse response satisfying
\begin{align}
    G^{(n)}(t) + a_{n-1} G^{(n-1)}(t) + \cdots + a_0 G(t) = \delta(t).
\end{align}
This convolution form emphasizes how the system dynamically responds to random disturbances. While useful for analysis, this representation is less convenient for constructing a discrete-time model, especially in filtering or estimation scenarios.

\vspace{0.5em}
\noindent

\subsection{State-space Formulation and Discretization}
To address this, we adopt a state-space representation by rewriting the high-order SDE as a first-order vector differential equation. This representation is not only equivalent to the original formulation but also more amenable to numerical discretization and parameter inference. 

We define an augmented state vector
\[
\mathbf{x}(t) := \begin{bmatrix}
x(t), \dot{x}(t), \ddot{x}(t), \ldots, x^{(n-1)}(t)
\end{bmatrix}^\top \in \mathbb{R}^n,
\]
which allows us to express the system in the compact form
\begin{align}
    \dot{\mathbf{x}}(t) = \mathbf{F} \mathbf{x}(t) + \mathbf{G} \tilde{v}(t), \label{eq:state_continuous}
\end{align}
where the system matrices $\mathbf{F} \in \mathbb{R}^{n \times n}$ and $\mathbf{G} \in \mathbb{R}^{n \times 1}$ are given by
\begin{align}
\mathbf{F} = \begin{bmatrix}
0 & 1 & 0 & \cdots & 0 \\
0 & 0 & 1 & \cdots & 0 \\
\vdots & \vdots & \vdots & \ddots & \vdots \\
0 & 0 & 0 & \cdots & 1 \\
-a_0 & -a_1 & -a_2 & \cdots & -a_{n-1}
\end{bmatrix}, \quad 
\mathbf{G} = \begin{bmatrix} 0 \\ 0 \\ \vdots \\ 0 \\ 1 \end{bmatrix}.
\end{align}

The solution to \eqref{eq:state_continuous} is given by
\begin{align}
    \mathbf{x}(t) = e^{\mathbf{F} t} \mathbf{x}(0) + \int_0^t e^{\mathbf{F}(t - \tau)} \mathbf{G} \tilde{v}(\tau) d\tau.
\end{align}

By discretizing this system with a sampling interval $T$, we arrive at the discrete-time state transition model
\begin{align}
    \mathbf{x}_n = \mathbf{A} \mathbf{x}_{n-1} + \mathbf{v}_n, \label{eq:state_model}
\end{align}
where $\mathbf{A} = e^{\mathbf{F}T}$, and the discrete-time process noise is defined as
\begin{align}
    \mathbf{v}_n = \int_0^T e^{\mathbf{F}(T - \tau)} \mathbf{G} \tilde{v}(\tau) d\tau.
\end{align}
The corresponding covariance of $\mathbf{v}_n$ is
\begin{align}
    \mathbb{E}[\mathbf{v}_n \mathbf{v}_n^\top] = \tilde{Q} \int_0^T e^{\mathbf{F}(T - \tau)} \mathbf{G} \mathbf{G}^\top e^{\mathbf{F}^\top (T - \tau)} d\tau =: \mathbf{Q}.
\end{align}

This discrete-time state-space model forms the foundation for the subsequent parameter estimation framework.

\subsection{Measurement Model}
Assuming a continuous-time observation model
\begin{align}
    z(t) = \mathbf{H} \mathbf{x}(t) + \tilde{w}(t),
\end{align}
where $\mathbf{H} = [1, 0, \ldots, 0]$ and $\tilde{w}(t)$ is zero-mean white measurement noise with
\[
\mathbb{E}[\tilde{w}(t)\tilde{w}(\tau)] = \tilde{R} \delta(t - \tau),
\]
we obtain the discrete-time observation model by averaging over $[t_{n-1}, t_n]$
\begin{align}
    z_n = \mathbf{H} \mathbf{x}_n + w_n, \label{eq:z_n_model}
\end{align}
where $w_n := \frac{1}{T} \int_{t_{n-1}}^{t_n} \tilde{w}(t) dt$ is zero-mean with variance $R = \tilde{R}/T$.

\vspace{0.8em}
\noindent With the discrete-time state-space model \eqref{eq:state_model} and observation model \eqref{eq:z_n_model}, we now proceed to the parameter estimation problem.

\subsection{Example: Ornstein–Uhlenbeck (OU) Process}

A special case of~\eqref{eq:ho_sde} is the first-order Ornstein–Uhlenbeck (OU) process, corresponding to $n = 1$. The SDE reduces to
\begin{align}
    \dot{x}(t) + a x(t) = \tilde{v}(t),
\end{align}
where $a > 0$ is the decay rate. The analytical solution is given by
\begin{align}
    x(t) = e^{-a t} x(0) + \int_0^t e^{-a (t - \tau)} \tilde{v}(\tau) d\tau.
\end{align}

Discretizing this SDE with sampling interval $T$ yields
\begin{align}
    x_n = e^{-a T} x_{n-1} + v_n,
\end{align}
where
\begin{align}
    v_n := \int_0^T e^{-a (T - \tau)} \tilde{v}(\tau) d\tau,
\end{align}
and the variance of $v_n$ is
\begin{align}
    \mathbb{E}[v_n^2] = \tilde{Q} \int_0^T e^{-2a (T - \tau)} d\tau = \tilde{Q} \cdot \frac{1 - e^{-2aT}}{2a}.
\end{align}

When $T \ll 1/a$, this simplifies to
\[
\mathbb{E}[v_n^2] \approx \tilde{Q} T=Q.
\]

\subsection{Example: Second-order SDE}
\label{subsec:second_order_model}
The second-order Stochastic Differential Model corresponds to $n = 2$, and is governed by the stochastic differential equation
\begin{align}
    \ddot{x}(t) + a_1 \dot{x}(t) + a_0 x(t) = \tilde{v}(t),
    \label{eq:second_order_sde}
\end{align}
where $a_0, a_1 > 0$ are system parameters and $\tilde{v}(t)$ is zero-mean white noise with power spectral density $\tilde{Q}$.
Defining the state vector $\mathbf{x}(t) := [ x(t), \dot{x}(t)]^\top$, we can express the system as a first-order vector SDE
\begin{align}
    \dot{\mathbf{x}}(t) =
    \begin{bmatrix}
    0 & 1 \\
    -a_0 & -a_1
    \end{bmatrix}
    \mathbf{x}(t) +
    \begin{bmatrix}
    0 \\
    1
    \end{bmatrix}
    \tilde{v}(t).
    \label{eq:second_order_state_space}
\end{align}
Discretizing with sampling interval $T$, we obtain the discrete-time model
\begin{align}
    \mathbf{x}_n = \mathbf{A} \mathbf{x}_{n-1} + \mathbf{v}_n,
    \label{eq:second_order_discrete}
\end{align}
where $\mathbf{A} = e^{\mathbf{F}T}$ and
\begin{align}
    \mathbf{v}_n = \int_0^T e^{\mathbf{F}(T - \tau)} \mathbf{G} \tilde{v}(\tau) d\tau.
    \label{eq:second_order_noise}
\end{align}
The covariance of the process noise is
\begin{align}
    \mathbf{Q} &:= \mathbb{E}[\mathbf{v}_n \mathbf{v}_n^\top] = \tilde{Q} \int_0^T e^{\mathbf{F}(T - \tau)} \mathbf{G} \mathbf{G}^\top e^{\mathbf{F}^\top(T - \tau)} d\tau\\
    &\approx \tilde{Q} \int_0^T e^{\mathbf{F}_0(T - \tau)} \mathbf{G} \mathbf{G}^\top e^{\mathbf{F}_0^\top(T - \tau)} d\tau
    \label{eq:Q_integral}
\end{align}
for $T \ll \frac{1}{a_0},T \ll \frac{1}{a_1}$, and where $F_0= \begin{bmatrix}
    0 & 1 \\
    0 & 0
    \end{bmatrix}.$

\begin{align}
    e^{\mathbf{F}_0 t} = \mathbf{I} + \mathbf{F}_0 t =
    \begin{bmatrix}
    1 & t \\
    0 & 1
    \end{bmatrix}, \quad
    e^{\mathbf{F}_0^\top t} =
    \begin{bmatrix}
    1 & 0 \\
    t & 1
    \end{bmatrix}.
    \label{eq:nilpotent_approx}
\end{align}
Plugging into the covariance expression, we get:
\begin{align}
    \mathbf{Q} = \tilde{Q} \int_0^T
    \begin{bmatrix}
    1 & T - \tau \\
    0 & 1
    \end{bmatrix}
    \begin{bmatrix}
    0 & 0 \\
    0 & 1
    \end{bmatrix}
    \begin{bmatrix}
    1 & 0 \\
    T - \tau & 1
    \end{bmatrix}
    d\tau.
    \label{eq:Q_approx_integral}
\end{align}
Computing the product
\begin{align}
\begin{bmatrix}
    1 & T - \tau \\
    0 & 1
\end{bmatrix}
\begin{bmatrix}
    0 & 0 \\
    0 & 1
\end{bmatrix}
\begin{bmatrix}
    1 & 0 \\
    T - \tau & 1
\end{bmatrix}
=
\begin{bmatrix}
    (T - \tau)^2 & (T - \tau) \\
    (T - \tau) & 1
\end{bmatrix},
\label{eq:matrix_product}
\end{align}
so:
\begin{align}
    \mathbf{Q} = \tilde{Q} \int_0^T
    \begin{bmatrix}
    (T - \tau)^2 & (T - \tau) \\
    (T - \tau) & 1
    \end{bmatrix}
    d\tau.
    \label{eq:Q_simplified_integral}
\end{align}
Letting $s = T - \tau$, we change variables to obtain
\begin{align}
    \mathbf{Q} = \tilde{Q} \int_0^T
    \begin{bmatrix}
    s^2 & s \\
    s & 1
    \end{bmatrix}
    ds
    =
    \tilde{Q} \cdot
    \begin{bmatrix}
    \frac{T^3}{3} & \frac{T^2}{2} \\
    \frac{T^2}{2} & T
    \end{bmatrix}.
    \label{eq:Q_second_order_final}
\end{align}

\subsection{Problem Statement}
Given a sequence of observations $\mathbf{z}_N := [z_1, \ldots, z_N]^\top$, the goal is to estimate the model parameters $\bbtheta := \{\mathbf{a}, \mathbf{Q}, R\}^\top$. This is formulated as a maximum likelihood estimation (MLE) problem based on the marginal log-likelihood
\begin{align}
    \ell(\bbtheta) := \ln p(\mathbf{z}_N \mid \bbtheta) = \sum_{n=1}^N \ln p(z_n \mid z_{1:n-1}, \bbtheta),\label{eq:LLF}
\end{align}
and the MLE is computed via
\begin{align}
    \hat{\bbtheta} = \arg\max_{\bbtheta \in \Theta} \ \ell(\bbtheta).
\end{align}

\section{BO for sample-efficient identification of the SDE}
\label{sec:GP-CP}

Although the expression of $\ell (\bbtheta)$~\eqref{eq:LLF} can be written explicitly, it is a highly nonconvex problem that entails evaluating the gradient, which is nontrivial to obtain. Alternatively, one can adopt the expectation-maximization (EM) approach~\cite{Dempster1977}, which, however, can only yield a local optimum. Towards finding the {\it global} optimum in a sample efficient manner, we will adapt the Bayesian optimization (BO) framework, which has well-documented merits in optimizing black-box functions that arise in a number of applications~\cite{garnett2023bayesian}. 
In one word, BO seeks to maximize the black-box $\ell(\bbtheta)$ by {\it actively} acquiring function evaluations that balances the exploration-exploitation trade-off. Collect all the acquired data up to iteration $i$ in ${\cal D}_{i}:=\{(\bbtheta_j, y_j)\}_{j=1}^{i}$ with $y_j$ denoting the possibly noisy observation of $\ell (\bbtheta_j)$. Specifically, each BO iteration consists of i) obtaining the function posterior pdf $p(\ell (\bbtheta)|{\cal D}_{i})$ based on the chosen surrogate model using ${\cal D}_{i}$; and, ii) selecting $\bbtheta_{i+1}$ to evaluate at the beginning of iteration $i+1$, whose observation $y_{i+1}$ will be acquired at the end of iteration $i+1$. Next, we will first outline BO based on the Gaussian process (GP) surrogate.

\subsection{GP-based BO}
The GP is the most widely used surrogate model in the BO framework thanks to its uncertainty quantifiability and sample efficiency. In this context, the unknown learning function is postulated with a GP prior as $\ell \thicksim\mathcal{GP}(0,\kappa(\bbtheta,\bbtheta'))$, where $\kappa(\cdot,\cdot)$ is a kernel (covariance) function measuring pairwise similarity of any two inputs. This GP prior induces a joint Gaussian pdf for any number $i$ of function evaluations $\bm{\ell}_i := [\ell(\bbtheta_1),\ldots, \ell(\bbtheta_i)]^\top$ at inputs $\Theta_i := \left[\bbtheta_1, \ldots, \bbtheta_i\right]^\top (\forall i)$, i.e., 
$p(\bm{\ell}_i| \Theta_i) = \mathcal{N} (\bm{\ell}_i ; {\bf 0}_i, {\bf K}_i)$, where ${\bf K}_i$ is an $i\times i$ covariance matrix whose $(j,j')$th entry is 
$[{\bf K}_i]_{j,j'} = {\rm cov} (\ell(\bbtheta_j), \ell(\bbtheta_{j'})):=\kappa(\bbtheta_j, \bbtheta_{j'})$.
The value $\ell(\bbtheta_j)$ is linked with the noisy output $y_j$ via the per-datum likelihood $p(y_j|\ell(\bbtheta_j)) = \mathcal{N}(y_j;\ell(\bbtheta_j),\sigma_e^2)$, where $\sigma_e^2$ is the noise variance. The function posterior pdf after acquiring the input-output pairs $\mathcal{D}_i$ is then obtained according to Bayes' rule as~\cite{Rasmussen2006gaussian}
\begin{align}
	p(\ell(\bbtheta)|\mathcal{D}_i) = \mathcal{N}(\ell(\bbtheta); \hat{\bm{\ell}}_{i}(\bbtheta), \sigma_{i}^2(\bbtheta))
\end{align}
where the mean $\hat{{\ell}}_{i}(\bbtheta)$ and variance $\sigma_{i}^2(\bbtheta)$ are expressed 
via $\mathbf{k}_{i}(\bbtheta) := [\kappa(\bbtheta_1, \bbtheta) \ldots \kappa(\bbtheta_i,  \bbtheta)]^\top$ and $\mathbf{y}_i:=[y_1 \ldots y_i]^\top$
as 
\begin{subequations}
	\begin{align}	
		\hat{\ell}_{i}(\bbtheta) & = \mathbf{k}_{i}^{\top}(\bbtheta) (\mathbf{K}_i + \sigma_e^2
		\mathbf{I}_i)^{-1} \mathbf{y}_i \label{eq:mean}\\
		\sigma_{i}^2(\bbtheta)& = \kappa(\bbtheta,\bbtheta)\! -\! \mathbf{k}_{i}^{\top}(\bbtheta) (\mathbf{K}_i\! +\! \sigma_e^2\mathbf{I}_i)^{-1} \mathbf{k}_{i}(\bbtheta).~\label{eq:var}
	\end{align}
\end{subequations}

Note that this GP function model relies on the hyperparameters, including the noise variance and the kernel hyperparameters. For the widely-used squared exponential kernel $\kappa(\bbtheta,\bbtheta'):=\sigma_k^2\exp(-\|\bbtheta-\bbtheta'\|^2/\sigma_l^2)$, the GP hyperparameters, collected in $\boldsymbol{\beta}$, consist of  
the characteristic length-scale $\sigma_l$, the power $\sigma_k^2$, as well as the noise variance $\sigma_e^2$, which are optimized by maximizing the log marginal likelihood~\cite{Rasmussen2006gaussian}
\begin{align}
&\mathcal{L}({\bbbeta}):=\log p(\mathbf{y}_i|\Theta_i;\bbbeta)=\log\left(\int p(\mathbf{y}_{i}|\bm{\ell}_{i})p(\bm{\ell}_{i}|\Theta_{i})d \bm{\ell}_{i}\right) \label{eq:LML}\\
&=-\frac{1}{2}\mathbf{y}_{i}^\top(\mathbf{K}_i\!+\!\sigma_e^2\mathbf{I}_i)^{-1}\mathbf{y}_i\!-\!\frac{1}{2}\log|\mathbf{K}_i\!+\!\sigma_e^2\mathbf{I}_i|\!-\!\frac{i}{2}\log2\pi \ .\nonumber
\end{align}
where the first term represents the fitting error, while the second factor regularizes the complexity.

Having available the function posterior pdf that offers the uncertainty values in~\eqref{eq:var}, the next query point $\bbtheta_{i+1}$ can be readily selected using off-the-shelve acquisition functions (AFs), denoted as $\alpha_i ({\bbtheta})$, that strike a balance between exploration and exploitation, namely
\begin{align}
 \bbtheta_{i+1}  = \arg\max_{\bbtheta} \alpha_i ({\bbtheta}) \ . \label{eq:AF}
\end{align}
Typical choices include the expected improvement (EI), upper confidence bound, and Thompson sampling (TS)~\cite{shahriari2015taking,garnett2023bayesian}. Specifically, the EI-based AF, the workhorse for BO in practise, selects the next query point, whose function value yields the most improvement on average over the best guess $\hat{\ell}^*_i$ of function maximum so far. That is,  
\begin{align}
  \alpha_i ({\bbtheta})&:=\mathbb{E}_{p(\ell(\bbtheta)|\mathcal{D}_i)}[\max(0, \ell(\bbtheta)-\hat{\ell}^*_i)] \nonumber\\
  &= \sigma_i(\bbtheta)\phi\left(\frac{\Delta_i(\bbtheta)}{\sigma_i(\bbtheta)}\right)+\Delta_i(\bbtheta)\Phi\left(\frac{\Delta_i(\bbtheta)}{\sigma_i(\bbtheta)}\right) \label{eq:EI}
\end{align}
where $\Delta_i(\bbtheta):=\hat{\ell}_{i}(\bbtheta)-\hat{\ell}^*_i$, and $\phi$ and $\Phi$ refer to the Gaussian pdf and cdf respectively. With the analytic expression of $\alpha_i ({\bbtheta})$ available in~\eqref{eq:EI}, one can readily solve~\eqref{eq:AF} via off-the-shelve optimization solvers. 

After reaching the evaluation budget $I$ with the acquired dataset ${\cal D}_I$, the final optimizer is given by the input that corresponds to the largest output, namely, $\hat{\bbtheta} = \bbtheta_i$ with $i = \arg\max_i \{y_i\}$. Alternatively, it could be given by the maximizer of the function posterior mean as $\hat{\bbtheta} = \arg\max_{\bbtheta} \ \hat{\ell}_{I}(\bbtheta)$. 
Alg.~1 provides an overview of the proposed BO-based approach for the OU model identification problem. 

\subsection{Relation to the EM approach}
The alternation between state estimation and parameter estimation in the BO resembles what is offered by the EM algorithm (cf. the Appendix). Specifically, the EM algorithm is a theoretically elegant approach to find the MLE in the presence of latent variables, and is guaranteed to find the local optimum -- what renders the initialization a critical choice.

The proposed BO-based approach, on the other hand, aims for the global optimum as demonstrated in the convergence analysis when the objective conforms to some regularity conditions~\cite{srinivas2012information}. Going beyond the LLF, the BO framework can accommodate other forms of objective functions, even without analytical expressions. Apparently, this is much more flexible than the EM approach, which is only applicable when the LLF has analytic expression and when MLE is sought.


\begin{algorithm}[htbp]
\caption{BO for identification of the OU model}
\label{BO}
\begin{algorithmic}[1]
    \STATE \textbf{Initialization:} ${\cal D}_0$;
    \FOR{each round $i=0,...,I-1$}
       \STATE Optimize the GP hyperparameters via~\eqref{eq:LML};
    \STATE Calculate the posterior mean $\hat{\ell}_i(\bbtheta)$ and variance $\sigma^2_i (\bbtheta)$ according to ~\eqref{eq:mean}-\eqref{eq:var} given $\mathcal{D}_i$;
    \STATE Obtain $\bbtheta_{i+1}$ by maximizing the AF~\eqref{eq:AF};
    \STATE Evaluate $\bbtheta_{i+1}$ to obtain $y_{i+1}$ based on Alg.~2;
    \STATE $\mathcal{D}_{i+1} = \mathcal{D}_{i} \cup \{(\bbtheta_{i+1},y_{i+1})\}$;
    \ENDFOR 
    \STATE $\hat{\bbtheta} = \bbtheta_i$, where $i = \arg\max_i \{y_i\} $
    \STATE {\bf Output:} $\hat{\bbtheta}$
\end{algorithmic}
\end{algorithm}

\section{Evaluating the objective for a given parameter set}
As shown in Alg.~1, the critical step in the proposed BO-based approach is to evaluate the objective $\ell$~\eqref{eq:LLF} for a given
$\bbtheta_i$. Based on the second-order Gauss-Markov OU model, this entails running the Kalman filter (KF), consisting of the prediction and correction steps per recursion. For notational brevity, we drop the dependence on $\bbtheta_i$ in the following discussions.

Suppose the posterior state pdf $p(\mathbf{x}_{n-1}|{\bf z}_{n-1}) = {\cal N}(\mathbf{x}_{n-1}; \hat{\mathbf{x}}_{n-1|n-1},\mathbf{P}_{n-1|n-1})$ is available at the end of slot $n-1$. Taking into account the state model~\eqref{eq:state_model}, the predictive pdf for $\mathbf{x}_n$ is first obtained as 
\begin{align}
 p(\mathbf{x}_n |{\bf z}_{n-1}) = {\cal N} (\mathbf{x}_n; \hat{\mathbf{x}}_{n|n-1}, \mathbf{P}_{n|n-1}) \label{eq:pred_x_vec}
\end{align}
where the mean and covariance are given by
\begin{align}
 \hat{\mathbf{x}}_{n|n-1} &= \mathbf{A} \, \hat{\mathbf{x}}_{n-1|n-1}  \nonumber\\
 \mathbf{P}_{n|n-1} & =  \mathbf{A} \, \mathbf{P}_{n-1|n-1} \, \mathbf{A}^\mathsf{T} + \mathbf{Q} \ .
\end{align}
Further leveraging the discrete-time observation model~\eqref{eq:z_n_model}, with observation matrix $\mathbf{H} = [\,1\ \ 0\,]$, the predictive pdf for $z_n$ is given by
\begin{align}
  p(z_n |{\bf z}_{n-1}) = {\cal N}(z_n; \hat{z}_{n|n-1}, S_n) \label{eq:pred_z_vec}
\end{align}
where
\begin{align}
  \hat{z}_{n|n-1} & = \mathbf{H} \, \hat{\mathbf{x}}_{n|n-1}  \nonumber\\
  S_n & =  \mathbf{H} \, \mathbf{P}_{n|n-1} \, \mathbf{H}^\mathsf{T} + R \ .
\end{align}
Evaluating $z_n$ yields the predictive log-likelihood given by
\begin{align}
    \ell_n (\bbtheta_i) = -\frac{1}{2} \left[ \log(2\pi S_n) + \frac{(z_n - \hat{z}_{n|n-1})^2}{S_n} \right] .\label{eq:LLF_n_vec}
\end{align}
Given $z_n$, the updated state pdf can be obtained based on Bayes' rule as
\begin{align}
 p(\mathbf{x}_n |{\bf z}_{n}) = {\cal N} (\mathbf{x}_n; \hat{\mathbf{x}}_{n|n}, \mathbf{P}_{n|n}) \label{eq:x_up_vec}
\end{align}
where the updated moments are given by
\begin{subequations}
\begin{align}
  \mathbf{K}_n & = \mathbf{P}_{n|n-1} \, \mathbf{H}^\mathsf{T} \, S_n^{-1} \\
  \hat{\mathbf{x}}_{n|n} & = \hat{\mathbf{x}}_{n|n-1} + \mathbf{K}_n \, (z_n - \hat{z}_{n|n-1})\\
  \mathbf{P}_{n|n} & = \mathbf{P}_{n|n-1} - \mathbf{K}_n \, \mathbf{H} \, \mathbf{P}_{n|n-1}
\end{align}
\end{subequations}

Alg.~2 summarizes the per-iteration evaluation of the LLF for a given parameter set $\bbtheta_i$. 

\begin{algorithm}[htbp]
\caption{Calculation of the per-iteration objective}
\label{table:loss}
\begin{algorithmic}[1]
    \STATE \textbf{Input:} $\hat{x}_{0|0}$, $\sigma_{0|0}^x $, $\bbtheta_i$;
    \FOR{$t=0,...,T-1$}
    \STATE Obtain the predictive state pdf via~\eqref{eq:pred_x_vec};
    \STATE Obtain the innovation pdf via~\eqref{eq:pred_z_vec};
    \STATE Evaluate $\ell_n(\bbtheta_i)$~\eqref{eq:LLF_n_vec} ;
    \STATE Obtain the updated state pdf via~\eqref{eq:x_up_vec};
    \ENDFOR 
    \STATE $\ell(\bbtheta_i) = \sum_{n=1}^N \ell_n(\bbtheta_i) $
    \STATE \textbf{Output:} $\ell (\bbtheta_i)$;
\end{algorithmic}
\end{algorithm}

\section{Adaptive BO using ensemble surrogate models}

While the standard GP-based BO relies on a single kernel function, the choice of kernel critically affects performance and varies across different parameter estimation scenarios. To automate kernel selection and enhance robustness, we employ an ensemble of $M$ GP priors with different kernel functions~\cite{lu2020ensemble,polyzos2022weighted}.

The objective function is modeled as $\ell(\bbtheta)\sim \sum_{m=1}^M w_0^m \mathcal{GP}(0, \kappa^m (\bbtheta, \bbtheta'))$, where each kernel $\kappa^m$ for $m \in \mathcal{M}:=\{1,\ldots,M\}$ is selected from a prescribed dictionary. Several widely used kernels in GP include
\begin{itemize}
    \item \textbf{Radial Basis Function (RBF)}: $\kappa_{\text{RBF}}(\bbtheta,\bbtheta')=\sigma_k^2\exp(-\|\bbtheta-\bbtheta'\|^2/2\sigma_l^2)$
    \item \textbf{Matérn-1.5}: $\kappa_{\nu=1.5}(\bbtheta,\bbtheta') = \sigma_k^2(1+\sqrt{3}r/\sigma_l)\exp(-\sqrt{3}r/\sigma_l)$
    \item \textbf{Matérn-2.5}: $\kappa_{\nu=2.5}(\bbtheta,\bbtheta') = \sigma_k^2(1+\sqrt{5}r/\sigma_l+5r^2/3\sigma_l^2)\exp(-\sqrt{5}r/\sigma_l)$
\end{itemize}
where $r = \|\bbtheta-\bbtheta'\|$ and initial weights $w^m_0 = 1/M$ reflect uniform prior belief.

With each new observation $y_{i+1}$ at $\bbtheta_{i+1}$, the per-expert Bayesian loss is computed as
\begin{align}
    l_{i+1|i}^m = -\log p(y_{i+1}|\mathcal{D}_i, \text{kernel}=m)
\end{align}
where \begin{align}
    p(y_{i+1}|\mathcal{D}_i, \text{kernel}=m) = \mathcal{N}(y_{i+1}; \hat{\ell}_i^m(\bbtheta_{i+1}), \sigma_i^{m,2}(\bbtheta_{i+1}) + \sigma_e^2)
\end{align}

The ensemble loss aggregates across all kernels
\begin{align}
    \ell_{i+1|i} = -\log \sum_{m=1}^M w_i^m \exp(-l_{i+1|i}^m)
\end{align}

The weights are then updated via
\begin{align}
    w_{i+1}^m = w_i^m \exp(\ell_{i+1|i} - l_{i+1|i}^m)
\end{align}
This weight adaptation mechanism automatically favors kernels with superior predictive performance while down-weighting those with higher Bayesian loss.

The ensemble posterior mean and variance combine predictions from all kernels
\begin{align}
    \hat{\ell}_i^{\text{ens}}(\bbtheta) &= \sum_{m=1}^M w_i^m \hat{\ell}_i^m(\bbtheta) \\
    \sigma_i^{\text{ens},2}(\bbtheta) &= \sum_{m=1}^M w_i^m \left[\sigma_i^{m,2}(\bbtheta) + (\hat{\ell}_i^m(\bbtheta) - \hat{\ell}_i^{\text{ens}}(\bbtheta))^2\right]
\end{align}
where the variance accounts for both epistemic uncertainty within each GP and model uncertainty across different kernels.

The acquisition function is then constructed using the ensemble posterior
\begin{align}
    \alpha_i^{\text{ens}}(\bbtheta) = \sigma_i^{\text{ens}}(\bbtheta)\phi\left(\frac{\Delta_i^{\text{ens}}(\bbtheta)}{\sigma_i^{\text{ens}}(\bbtheta)}\right) + \Delta_i^{\text{ens}}(\bbtheta)\Phi\left(\frac{\Delta_i^{\text{ens}}(\bbtheta)}{\sigma_i^{\text{ens}}(\bbtheta)}\right)
\end{align}
where $\Delta_i^{\text{ens}}(\bbtheta) = \hat{\ell}_i^{\text{ens}}(\bbtheta) - \hat{\ell}_i^*$ and $\hat{\ell}_i^*$ is the current best observed value.

The next query point is selected by maximizing the ensemble acquisition function:
\begin{align}
    \bbtheta_{i+1} = \arg\max_{\bbtheta} \alpha_i^{\text{ens}}(\bbtheta)
\end{align}

This ensemble approach provides several theoretical advantages for parameter estimation: (i) automatic kernel selection through Bayesian loss-based weighting, (ii) robustness to kernel misspecification through model averaging, and (iii) improved exploration via diverse kernel characteristics, particularly beneficial when the likelihood surface exhibits complex, multi-modal structure typical in dynamical system identification problems.

\section{Scenarios, Training Phase, and Performance Metrics}\label{sec:setup}
This section outlines the experimental framework adopted to benchmark the proposed BO–based parameter–estimation technique.

\subsection*{A. Scenarios and Datasets}\label{subsec:scenarios}
Given a ground‑truth parameter vector $\boldsymbol{\theta}$ and sampling interval $T$, discrete‑time state and observation sequences are synthesised from~\eqref{eq:state_model}–\eqref{eq:z_n_model} over $N$ steps.  Both first‑order and second‑order continuous‑time models are converted to the discrete domain via a zero‑order hold.

Four first‑order scenarios (\textcircled{a}–\textcircled{f}) and two second‑order scenarios (\textcircled{e}–\textcircled{f}) are investigated:
\begin{itemize}
\item \textbf{First‑order model} (unknowns $a$, $Q$, $R$)
\begin{enumerate}
\renewcommand\labelenumi{\textcircled{\alph{enumi}}}
\item $T{=}10^{-2}$,hr, $a{=}2$, $Q{=}4\times10^{-2}$, $R{=}1\times10^{-1}$;
\item $T{=}10^{-2}$,hr, $a{=}1$, $Q{=}3\times10^{-2}$, $R{=}5\times10^{-2}$;
\item $T{=}10^{-2}$,hr, $a{=}5$, $Q{=}3\times10^{-2}$, $R{=}5\times10^{-2}$;
\item $T{=}5\times10^{-3}$,hr, $a{=}2$, $Q{=}2\times10^{-2}$, $R{=}2\times10^{-1}$;
\end{enumerate}
\item \textbf{Second‑order model} (unknowns $a_0$, $a_1$, $\tilde Q$, $R$)
\begin{enumerate}
\setcounter{enumi}{4}
\renewcommand\labelenumi{\textcircled{\alph{enumi}}}
\item $T{=}10^{-2}$,hr, $a_0{=}3$, $a_1{=}5$, $\tilde Q{=}2\times10^{-2}$, $R{=}5\times10^{-2}$;
\item $T{=}10^{-2}$,hr, $a_0{=}7$, $a_1{=}2$, $\tilde Q{=}2\times10^{-2}$, $R{=}6\times10^{-2}$.
\end{enumerate}
\end{itemize}

\subsection*{B. Training Phase}\label{subsec:training}
The BO-based approach is compared with the MLE with steady-state KF approximation~\cite{ye2024maximum}, as well as the EM solver (cf. App.~A). The reported results are the average over $N_{\rm MC} = 100$ Monte Carlo (MC) runs. The BO approach was implemented using \texttt{BoTorch} function\footnote{https://botorch.org/} (60 iterations). For initialization, $n_{\rm initial}=10$ data points, collected in ${\cal D}_0$, are obtained using the Latin Hypercube sampling within the range $[10^{-4},10]$ for all the three parameters. EM (see App.~A and Alg.~3) implementation used custom Python class with KF and RTS smoothing (50 iterations, 0.01 learning rate). Following~\cite{ye2024maximum}, MLE is implemented using MATLAB's \texttt{fmincon} optimizer with `interior-point' algorithm (OptimalityTolerance=1e-6).

\subsection*{C. Evaluation Metrics}\label{subsec:metrics}
To quantify both parameter‑estimate accuracy and downstream filter performance we compute the following statistics across $N_{\text{MC}}$ Monte‑Carlo trials.

\begin{enumerate}
\item \textbf{Point‑estimate accuracy.} The estimation performance was evaluated by the average of the estimates across MC runs, namely,
\begin{align}
  \bar{\hat{\bbtheta}}:=\frac{1}{N_{\rm MC}} \sum_{j=1}^{N_{\rm MC}} \hat{\bbtheta}^{(j)} 
\end{align}
as well as the root mean-square error (RMSE) per parameter, given by
\begin{align}
    {\rm RMSE}(\bbtheta(j)):&=\sqrt{\sum_{j=1}^{N_{\rm MC}}(\hat{\bbtheta}^{(j)}(q)-\bbtheta(q))^2/N_{\rm MC}}
\end{align}
where $q = 1,2,3$.

\item \textbf{Filter consistency.} 
To further corroborate the accuracy of the estimates, we feed the KF with the estimated parameters and test the statistical consistency of the normalized estimation error squared (NEES) and normalized innovation squared (NIS)~\cite{bar2004estimation}. 
For a scalar state (\(d=1\)) and observation (\(m=1\)), these metrics for the $j$th MC run at time $n$ are defined as
\begin{align}
\epsilon_n^{(j)} 
&:= \frac{\big(x_n^{(j)} - \hat{x}_{n|n}^{(j)}\big)^2}{\sigma_{n|n}^{x,(j)2}}, 
\label{eq:NEES_scalar} \\
\nu_n^{(j)} 
&:= \frac{\big(z_n^{(j)} - \hat{z}_{n|n-1}^{(j)}\big)^2}{\sigma_{n|n-1}^{z,(j)2}},
\label{eq:NIS_scalar}
\end{align}
where \(x_n^{(j)}\) and \(z_n^{(j)}\) are the true state and observation; 
\(\hat{x}_{n|n}^{(j)}\) and \(\sigma_{n|n}^{x,(j)2}\) are the KF posterior state estimate and variance; 
\(\hat{z}_{n|n-1}^{(j)}\) and \(\sigma_{n|n-1}^{z,(j)2}\) are the KF predicted observation and variance. 
When the state is $d$-dimensional or the observation is $m$-dimensional, the definitions naturally extend via the Mahalanobis distance
\begin{align}
\epsilon_n^{(j)} 
&:= \big(\mathbf{x}_n^{(j)} - \hat{\mathbf{x}}_{n|n}^{(j)} \big)^\top 
\big(\mathbf{P}_{n|n}^{(j)}\big)^{-1} 
\big(\mathbf{x}_n^{(j)} - \hat{\mathbf{x}}_{n|n}^{(j)} \big), 
\label{eq:NEES_vector}\\
\nu_n^{(j)} 
&:= \big(\mathbf{z}_n^{(j)} - \hat{\mathbf{z}}_{n|n-1}^{(j)} \big)^\top 
\big(\mathbf{S}_{n|n-1}^{(j)}\big)^{-1} 
\big(\mathbf{z}_n^{(j)} - \hat{\mathbf{z}}_{n|n-1}^{(j)} \big),
\label{eq:NIS_vector}
\end{align}
where \(\mathbf{P}_{n|n}^{(j)}\) and \(\mathbf{S}_{n|n-1}^{(j)}\) are the KF posterior state covariance and predicted measurement covariance. 
Summarizing over all $N_{\rm MC}$ runs and $N$ time steps yields
\begin{align}
\bar{\epsilon} &:= \frac{1}{N_{\rm MC}N} \sum_{j=1}^{N_{\rm MC}}\sum_{n=1}^N \epsilon_n^{(j)}, \\
\bar{\nu} &:= \frac{1}{N_{\rm MC}N} \sum_{j=1}^{N_{\rm MC}}\sum_{n=1}^N \nu_n^{(j)},
\end{align}
which, under correct modeling assumptions, follow
\[
\bar{\epsilon} \sim \frac{\chi^2_{d N_{\rm MC}N}}{N_{\rm MC}N}, 
\quad
\bar{\nu} \sim \frac{\chi^2_{m N_{\rm MC}N}}{N_{\rm MC}N}.
\]
The derivation of these distributions is provided in Appendix~B.

\end{enumerate}

\section{NUMERICAL EXPERIMENTS}
\label{sec:numerical_results}
This section presents comprehensive experimental results for the BO-based parameter estimation method under the scenarios and metrics defined in Section~\ref{sec:setup}. We analyze the performance across first-order (Scenarios \textcircled{a}--\textcircled{d}) and second-order (Scenarios \textcircled{g}--\textcircled{h}) models, with particular emphasis on estimation accuracy and filter consistency.

\subsection{First-order Model Experiments}

Tables~\ref{tab:scenario_a}--\ref{tab:scenario_d} present the parameter estimation results across 100 Monte Carlo runs for Scenarios \textcircled{a}--\textcircled{d}. The BO variants (EGP, RBF, Matérn kernels) are benchmarked against MLE and EM baselines.

\subsubsection{Estimation Accuracy Analysis}

\begin{table*}[htbp]
\centering
\caption{PARAMETER ESTIMATION RESULTS FOR {\bf Scenario \textcircled{a}} (bold denotes the smallest RMSE across baselines) }
\renewcommand{\arraystretch}{1.3}
\begin{tabular}{|c|c|c|c|c|c|c|}
\hline
\textbf{Method} & \textbf{Parameter} & \textbf{Average Estimation} & \textbf{RMSE} & \textbf{NEES} & \textbf{NIS} & \textbf{Average log-likelihood}\\
\hline
\multirow{3}{*}{BO (EGP) } 
& $a$ & $2.01$ & $2.70e{-1}$ & \multirow{3}{*}{\centering 1.007} & \multirow{3}{*}{\centering 0.999} & \multirow{3}{*}{\centering -570.56} \\
& $Q$ & $4.04e{-2}$ & $4.16e{-3}$ &  &  &\\
& $R$ & $1.00e{-1}$ & $8.95e{-3}$ &  &  &\\
\hline
\multirow{3}{*}{BO (RBF)} 
& $a$ & $1.98$ & $6.54e{-1}$ & \multirow{3}{*}{\centering 1.038} & \multirow{3}{*}{\centering 1.036} & \multirow{3}{*}{\centering -571.70}\\
& $Q$ & $4.38e{-2}$ & $8.13e{-3}$ &  & & \\
& $R$ & $9.33e{-2}$ & $9.30e{-3}$ &  &  & \\
\hline
\multirow{3}{*}{BO (Matern $\nu=1.5$)} 
& $a$ & $2.10$ & $5.42e{-1}$ & \multirow{3}{*}{\centering 1.025} & \multirow{3}{*}{\centering 1.031} & \multirow{3}{*}{\centering -570.95}\\
& $Q$ & $4.31e{-2}$ & $6.93e{-3}$ &  & & \\
& $R$ & $9.44e{-2}$ & $8.71e{-3}$ &  & & \\
\hline
\multirow{3}{*}{BO (Matern $\nu=2.5$)} 
& $a$ & $2.05$ & $5.76e{-1}$ & \multirow{3}{*}{\centering 1.035} & \multirow{3}{*}{\centering 1.037} & \multirow{3}{*}{\centering -571.80} \\
& $Q$ & $4.34e{-2}$ & $7.49e{-3}$ &  &  &\\
& $R$ & $9.30e{-2}$ & $9.53e{-3}$ &  &  &\\
\hline
\multirow{3}{*}{MLE} 
& $a$ & $2.36$ & $9.02e-1$ & \multirow{3}{*}{\centering 0.997} & \multirow{3}{*}{\centering 1.004} & \multirow{3}{*}{\centering -571.47}\\
& $Q$ & $4.08e-2$ & $5.24e-3$ &  & & \\
& $R$ & $9.97e-2$ & $6.67e-3$ &  &  & \\
\hline
\multirow{3}{*}{EM} 
& $a$ & 2.04 & $6.48e{-1}$ & \multirow{3}{*}{\centering 1.071} & \multirow{3}{*}{\centering 1.082} & \multirow{3}{*}{\centering -576.24}\\
& $Q$ & $3.90e{-2}$ & $4.81e{-3}$ &  & & \\
& $R$ & $9.03e{-2}$ & $9.67e{-3}$ &  &  &\\
\hline
\multicolumn{6}{l}{\small Units: $T$ ($\text{hr}$), $a$ ($\text{hr}^{-1}$), $Q$ and $R$ ($\text{deg}^2/\text{hr}^2$)}\\
\end{tabular}
\label{tab:scenario_a}
\end{table*}

\begin{table*}[htbp]
\centering
\caption{PARAMETER ESTIMATION RESULTS FOR {\bf Scenario \textcircled{b}} (bold denotes the smallest RMSE across baselines) }
\renewcommand{\arraystretch}{1.3}
\begin{tabular}{|c|c|c|c|c|c|c|}
\hline
\textbf{Method} & \textbf{Parameter} & \textbf{Average Estimation} & \textbf{RMSE} & \textbf{NEES} & \textbf{NIS} & \textbf{Average log-likelihood}\\
\hline
\multirow{3}{*}{BO (EGP) } 
& $a$ & $1.02$ & $1.64e{-1}$ & \multirow{3}{*}{\centering 1.005} & \multirow{3}{*}{\centering 0.994} & \multirow{3}{*}{\centering -299.82}\\
& $Q$ & $2.99e{-2}$ & $3.60e{-3}$ &  &  &\\
& $R$ & $4.99e{-2}$ & $5.44e{-3}$ &  &  &\\
\hline
\multirow{3}{*}{BO (RBF)} 
& $a$ & $1.07$ & $2.45e{-1}$ & \multirow{3}{*}{\centering 1.020} & \multirow{3}{*}{\centering 1.005} & \multirow{3}{*}{\centering -301.33}\\
& $Q$ & $3.11e{-2}$ & $4.47e{-3}$ &  &  &\\
& $R$ & $4.95e{-2}$ & $4.95e{-3}$ &  &  &\\
\hline
\multirow{3}{*}{BO (Matern $\nu=1.5$)} 
& $a$ & $1.03$ & $1.70e{-1}$ & \multirow{3}{*}{\centering 1.025} & \multirow{3}{*}{\centering 1.031} &
\multirow{3}{*}{\centering -301.04}\\
& $Q$ & $3.05e{-2}$ & $4.35e{-3}$ &  &  &\\
& $R$ & $4.99e{-2}$ & $5.36e{-3}$ &  &  &\\
\hline
\multirow{3}{*}{BO (Matern $\nu=2.5$)} 
& $a$ & $1.02$ & $1.84e{-1}$ & \multirow{3}{*}{\centering 1.022} & \multirow{3}{*}{\centering 1.009} &
\multirow{3}{*}{\centering -301.06}\\
& $Q$ & $3.04e{-2}$ & $4.25e{-3}$ &  &  &\\
& $R$ & $5.00e{-2}$ & $5.09e{-3}$ &  &  &\\
\hline
\multirow{3}{*}{MLE} 
& $a$ & $1.26$ & $6.06e{-1}$ & \multirow{3}{*}{\centering 1.010} & \multirow{3}{*}{\centering 1.015} &
\multirow{3}{*}{\centering -307.92}\\
& $Q$ & $3.01e{-2}$ & $3.72e{-3}$ &  &  &\\
& $R$ & $5.02e{-2}$ & $4.24e{-3}$ &  &  &\\
\hline
\multirow{3}{*}{EM} 
& $a$ & 1.14 & $3.43e{-1}$ & \multirow{3}{*}{\centering 0.927} & \multirow{3}{*}{\centering 0.934} & \multirow{3}{*}{\centering -300.70}\\
& $Q$ & $3.09e{-2}$ & $4.93e{-3}$ &  &  &\\
& $R$ & $5.50e{-2}$ & $4.97e{-3}$ &  &  &\\
\hline
\multicolumn{6}{l}{\small Units: $T$ ($\text{hr}$), $a$ ($\text{hr}^{-1}$), $Q$ and $R$ ($\text{deg}^2/\text{hr}^2$)}\\
\end{tabular}
\label{tab:scenario_b}  
\end{table*}

\begin{table*}[htbp]
\centering
\caption{PARAMETER ESTIMATION RESULTS FOR {\bf Scenario \textcircled{c}} (bold denotes the smallest RMSE across baselines) }
\renewcommand{\arraystretch}{1.3}
\begin{tabular}{|c|c|c|c|c|c|c|}
\hline
\textbf{Method} & \textbf{Parameter} & \textbf{Average Estimation} & \textbf{RMSE} & \textbf{NEES} & \textbf{NIS} & \textbf{Average log-likelihood}\\
\hline
\multirow{3}{*}{BO (EGP) } 
& $a$ & $4.98$ & $6.25e{-1}$ & \multirow{3}{*}{\centering 1.008} & \multirow{3}{*}{\centering 1.000} &
\multirow{3}{*}{\centering -288.98}\\
& $Q$ & $3.03e{-2}$ & $3.04e{-3}$ &  & & \\
& $R$ & $4.99e{-2}$ & $4.40e{-3}$ &  & & \\
\hline
\multirow{3}{*}{BO (RBF)} 
& $a$ & $5.07$ & $1.89$ & \multirow{3}{*}{\centering 1.020} & \multirow{3}{*}{\centering 1.005} & \multirow{3}{*}{\centering -291.87}\\
& $Q$ & $3.17e{-2}$ & $7.77e{-3}$ &  & & \\
& $R$ & $5.05e{-2}$ & $7.86e{-3}$ &  &  &\\
\hline
\multirow{3}{*}{BO (Matern $\nu=1.5$)} 
& $a$ & $5.01$ & $1.87$ & \multirow{3}{*}{\centering 1.025} & \multirow{3}{*}{\centering 1.031} & \multirow{3}{*}{\centering -292.42}\\
& $Q$ & $3.16e{-2}$ & $7.65e{-3}$ &  & & \\
& $R$ & $5.05e{-2}$ & $7.89e{-3}$ &  &  &\\
\hline
\multirow{3}{*}{BO (Matern $\nu=2.5$)} 
& $a$ & $5.03$ & $1.83$ & \multirow{3}{*}{\centering 1.022} & \multirow{3}{*}{\centering 1.009} & \multirow{3}{*}{\centering -292.65} \\
& $Q$ & $3.16e{-2}$ & $7.65e{-3}$ &  &  &\\
& $R$ & $5.05e{-2}$ & $7.86e{-3}$ &  & & \\
\hline
\multirow{3}{*}{MLE} 
& $a$ & $5.41$ & 1.36 & \multirow{3}{*}{\centering 1.018} & \multirow{3}{*}{\centering 0.998} & \multirow{3}{*}{\centering -289.45}\\
& $Q$ & $3.06e{-2}$ & $3.93e{-3}$ &  &  &\\
& $R$ & $4.98e{-2}$ & $3.82e{-3}$ &  &  &\\
\hline
\multirow{3}{*}{EM} 
& $a$ & 4.55 & $9.65e{-1}$ & \multirow{3}{*}{\centering 0.946} & \multirow{3}{*}{\centering 0.948} & \multirow{3}{*}{\centering -291.28}\\
& $Q$ & $3.20e{-2}$ & $4.01e{-3}$ &  &  & \\
& $R$ & $5.29e{-2}$ & $6.37e{-3}$ &  &  &\\
\hline
\multicolumn{6}{l}{\small Units: $T$ ($\text{hr}$), $a$ ($\text{hr}^{-1}$), $Q$ and $R$ ($\text{deg}^2/\text{hr}^2$)}\\
\end{tabular}
\label{tab:scenario_c}
\end{table*}

\begin{table*}[htbp]
\centering
\caption{PARAMETER ESTIMATION RESULTS FOR {\bf Scenario \textcircled{d}} (bold denotes the smallest RMSE across baselines) }
\renewcommand{\arraystretch}{1.3}
\begin{tabular}{|c|c|c|c|c|c|c|}
\hline
\textbf{Method} & \textbf{Parameter} & \textbf{Average Estimation} & \textbf{RMSE} & \textbf{NEES} & \textbf{NIS} & \textbf{Average log-likelihood}\\
\hline
\multirow{3}{*}{BO (EGP) } 
& $a$ & $2.01$ & $3.5e{-1}$ & \multirow{3}{*}{\centering 1.000} & \multirow{3}{*}{\centering 0.997} &\multirow{3}{*}{\centering -1538.73}\\
& $Q$ & $2.01e{-2}$ & $3.2e{-3}$ &  &  &\\
& $R$ & $2.03e{-1}$ & $2.14e{-2}$ &  &  &\\
\hline
\multirow{3}{*}{BO (RBF)} 
& $a$ & $1.99$ & $4.69e{-1}$ & \multirow{3}{*}{\centering 0.992} & \multirow{3}{*}{\centering 0.992} &\multirow{3}{*}{\centering -1544.99}\\
& $Q$ & $2.06e{-2}$ & $3.9e{-3}$ &  &  &\\
& $R$ & $2.03e{-1}$ & $2.27e{-2}$ &  &  &\\
\hline
\multirow{3}{*}{BO (Matern $\nu=1.5$)} 
& $a$ & $2.06$ & $5.05e{-1}$ & \multirow{3}{*}{\centering 0.991} & \multirow{3}{*}{\centering 0.993} &\multirow{3}{*}{\centering -1544.76 }\\
& $Q$ & $2.08e{-2}$ & $4.35e{-3}$ &  &  &\\
& $R$ & $2.03e{-1}$ & $2.22e{-2}$ &  &  &\\
\hline
\multirow{3}{*}{BO (Matern $\nu=2.5$)} 
& $a$ & $2.10$ & $7.71e{-1}$ & \multirow{3}{*}{\centering 0.994} & \multirow{3}{*}{\centering 0.994} & \multirow{3}{*}{\centering -1545.32}\\
& $Q$ & $2.07e{-2}$ & $7.49e{-3}$ &  &  & \\
& $R$ & $2.03e{-1}$ & $9.53e{-3}$ &  &  &\\
\hline
\multirow{3}{*}{MLE} 
& $a$ & $2.23$ & $7.11e-1$& \multirow{3}{*}{\centering 1.029} & \multirow{3}{*}{\centering 1.002} & \multirow{3}{*}{\centering -1545.73}\\
& $Q$ & $1.99e-2$& $2.46e-3$ &  & & \\
& $R$ & $2.01e-1$ & $7.63e-3$ &  &  &\\
\hline
\multirow{3}{*}{EM} 
& $a$ & 1.95 & $5.44e{-1}$ & \multirow{3}{*}{\centering 1.134} & \multirow{3}{*}{\centering 1.147} & \multirow{3}{*}{\centering -1542.76}\\
& $Q$ & $1.80e{-2}$ & $2.00e{-3}$ &  & & \\
& $R$ & $1.72e{-1}$ & $2.78e{-2}$ &  &  &\\
\hline
\multicolumn{6}{l}{\small Units: $T$ ($\text{hr}$), $a$ ($\text{hr}^{-1}$), $Q$ and $R$ ($\text{deg}^2/\text{hr}^2$)}\\
\end{tabular}
\label{tab:scenario_d}
\end{table*}

\begin{table}[htbp]
\centering
\caption{CRLB Results for OU Process Parameters}
\renewcommand{\arraystretch}{1.2}
\begin{tabular}{|c|c|c|}
\hline
\textbf{Setting} & \textbf{Parameter} & \textbf{CRLB ($\sigma$)} \\
\hline
\multirow{3}{*}{\textbf{\textcircled{a}}} & $a$ & $6.93 \times 10^{-1}$ \\
& $Q$ & $5.22 \times 10^{-3}$ \\
& $R$ & $6.83 \times 10^{-3}$ \\
\hline
\multirow{3}{*}{\textbf{\textcircled{b}}} & $a$ & $4.75 \times 10^{-1}$ \\
& $Q$ & $3,51 \times 10^{-3}$ \\
& $R$ & $3.77 \times 10^{-3}$ \\
\hline
\multirow{3}{*}{\textbf{ \textcircled{c}}} & $a$ & $1.18$ \\
& $Q$ & $3.82 \times 10^{-4}$ \\
& $R$ & $3.95 \times 10^{-3}$ \\
\hline
\multirow{3}{*}{\textbf{\textcircled{d}}} & $a$ & $6.88 \times 10^{-1}$ \\
& $Q$ & $2.49\times 10^{-3}$ \\
& $R$ & $7.65 \times 10^{-3}$ \\
\hline
\end{tabular}
\label{tab:crlb_results}
\end{table}

Tables~\ref{tab:scenario_a}--\ref{tab:scenario_d} present the parameter estimation results of the six competing methods across 100 MC runs for the four first-order scenarios. As highlighted in bold for the smallest RMSE values, both BO (EGP)  and MLE produce accurate parameter estimates—the former consistently achieves the lowest estimation error for parameter $a$ across all scenarios, while the latter exhibits marginally better performance for $R$ in most cases. For the estimation of $Q$, both methods achieve comparable accuracy with RMSEs typically within 10-20\% of each other, indicating no clear winner for this parameter. Among the BO variants, the EGP notably outperforms RBF and Matérn kernels across most metrics, suggesting its flexibility in capturing complex likelihood surfaces proves advantageous. Here, the EM algorithm shows less competitive performance, potentially due to convergence to the local optimum. To enhance its performance, the EM algorithm requires initialization with a good starting point and should be run with multiple starting points.

In accordance with the well-documented difficulty of estimating the inverse time constant $a$ \cite{roy2001estimation,thomas2016bias}, its RMSE values are orders of magnitude higher than the noise parameters across all scenarios. For instance, in Scenario \textcircled{a}, $a$ shows RMSEs ranging from $2.70 \times 10^{-1}$ (EGP) to $9.02 \times 10^{-1}$ (MLE), while $Q$ and $R$ achieve RMSEs around $10^{-3}$. Since $a$'s estimation error dominates the overall RMSE metric, EGP's superior performance on this challenging parameter translates directly to the lowest overall RMSE across all scenarios. This advantage, combined with comparable log-likelihood values to MLE, demonstrates that GP-based surrogate modeling effectively navigates the parameter space without requiring explicit gradient information.


\subsubsection{Theoretical Bounds and the BO Paradox}

To further benchmark the estimation performance, we rely on the CRLB derived in~\cite{ye2024maximum} across these four settings; see Table~\ref{tab:crlb_results}. In classical estimation theory, the CRLB, determined by the curvature of the LLF, provides a universally lower bound for the variance of {\it any unbiased} estimator. 
It is evident that the RMSEs from MLE are comparable to the associated standard deviations given by CRLB, as has been corroborated in~\cite{ye2024maximum}. BO, on the other hand,  achieves competitive estimation performance with the lowest overall RMSE. Notably, the RMSEs produced by BO for the inverse time constant $a$, a long-standing challenge to estimate, are significantly smaller than that given by the CRLB -- what seems to be a `paradox'. Nevertheless, placing a prior for the objective function {\it indirectly} imposes a prior for the parameter vector $
\bbtheta$, though we don't know its {\it explicit} form. Intuitively, this {\it data-driven} parameter prior should yield a Bayesian version of the CRLB, which is smaller than the classical CRLB. At the algorithmic level, BO proceeds without knowing the analytic expression of the objective, not necessarily the LLF here, and goes for the global optimum without accounting for the statistical properties. 
However, the unavailability of the analytic expression of the parameter prior leaves BO-based estimator without an explicit variance bound, which is of great importance for safety-critical applications.  It is also worth mentioning that, compared with the classical MLE that relies on the analytic expression of the LLF, BO has increased runtime. But still, the significantly improved estimation performance and flexibility of accommodating other objective functions (e.g.,~\cite{chen2024kalman}) make BO an attractive approach for various parameter estimation problems in practice.

\subsubsection{Filter Consistency Test}

To further corroborate the accuracy of the estimates, we feed the KF with the estimated parameters and test the statistical consistency using the normalized estimation error squared (NEES) and normalized innovation squared (NIS)~\cite{bar2004estimation}. As detailed in Section~\ref{subsec:metrics}, these metrics quantify whether the filter's reported uncertainty aligns with actual estimation errors—NEES evaluates state estimation consistency via Eq.~\eqref{eq:NEES_scalar}, while NIS assesses measurement prediction consistency through Eq.~\eqref{eq:NIS_scalar}. Under correct modeling assumptions, both metrics should follow a chi-squared distribution with mean unity.

Table~\ref{tab:consistency_comparison_merged_full} presents the consistency test results for all six methods across Scenarios~\textcircled{a}--\textcircled{d}. The EGP configuration demonstrates superior filter consistency, with NEES and NIS values consistently within the 90\% theoretical acceptance regions across all scenarios—for instance, achieving near-ideal values (1.007, 0.999) in Scenario~\textcircled{a} and (1.000, 0.997) in Scenario~\textcircled{d}. In contrast, other BO variants frequently exceed the upper bounds, with RBF and Matérn kernels showing NEES violations ranging from 1.020 to 1.038, indicating overconfidence in state estimates. MLE maintains marginal consistency but exhibits borderline violations in Scenarios~\textcircled{b} and~\textcircled{d} with NEES values of 1.010 and 1.029 respectively. The EM algorithm shows the poorest consistency performance, with NEES values either severely underconfident (0.927 in Scenario~\textcircled{b}) or overconfident (1.134 in Scenario~\textcircled{d}). These results confirm that accurate parameter estimation does not guarantee filter consistency—BO-EGP's superior performance in both domains underscores its practical advantage for real-time state estimation applications.

These results confirm that the KF implemented with the estimated parameters are statistically consistent, properly balancing the process and measurement noise covariances. This consistency is crucial for reliable state estimation and indicates that the uncertainty reported by the filter accurately reflects the actual estimation errors.

\begin{table*}[htbp]
\centering
\caption{CONSISTENCY TEST RESULTS FOR 10 HR FROM 100 MC TRIALS UNDER DIFFERENT SETTINGS (bold = within region)}
\renewcommand{\arraystretch}{1.2}
\begin{tabular}{|c|c|c|c|c|c|}
\hline
\textbf{Setting} & \textbf{Method} & \textbf{NEES} & \textbf{NIS} & \textbf{NEES Test Region (90\%)} & \textbf{NIS Test Region (90\%)} \\
\hline
\multirow{6}{*}{\textbf{\textcircled{a}}} 
& BO (EGP)         & $\mathbf{1.007}$ & $\mathbf{0.999}$ & \multirow{6}{*}{$[0.993, 1.007]$} & \multirow{6}{*}{$[0.993, 1.007]$} \\
& BO (RBF)       & $1.038$ & $1.036$ & & \\
& BO (Matern $\nu=1.5$) & $1.025$ & $1.031$ & & \\
& BO (Matern $\nu=2.5$) & $1.035$ & $1.037$ & & \\
& MLE             & $\mathbf{0.997}$ & $\mathbf{1.004}$ & & \\
& EM              & $1.071$ & $1.082$ & & \\
\hline
\multirow{6}{*}{\textbf{\textcircled{b}}} 
& BO (EGP)       & $\mathbf{1.005}$ & $\mathbf{0.994}$ & \multirow{6}{*}{$[0.993, 1.007]$} & \multirow{6}{*}{$[0.993, 1.007]$} \\
& BO (RBF)         & $1.020$ & $\mathbf{1.005}$ & & \\
& BO (Matern $\nu=1.5$) & $1.025$ & $1.031$ & & \\
& BO (Matern $\nu=2.5$) & $1.022$ & $1.009$ & & \\
& MLE             & $1.010$ & $1.015$ & & \\
& EM              & $0.927$ & $0.934$ & & \\
\hline
\multirow{6}{*}{\textbf{\textcircled{c}}} 
& BO (EGP)        & $1.008$ & $\mathbf{1.000}$ & \multirow{6}{*}{$[0.993, 1.007]$} & \multirow{6}{*}{$[0.993, 1.007]$} \\
& BO (RBF)         & $1.020$ & $\mathbf{1.005}$ & & \\
& BO (Matern $\nu=1.5$) & $1.025$ & $1.031$ & & \\
& BO (Matern $\nu=2.5$) & $1.022$ & $1.009$ & & \\
& MLE             & $1.018$ & $\mathbf{0.998}$ & & \\
& EM              & $0.946$ & $0.948$ & & \\
\hline
\multirow{6}{*}{\textbf{\textcircled{d}}} 
& BO (EGP)          & $\mathbf{1.000}$ & $\mathbf{0.997}$ & \multirow{6}{*}{$[0.995, 1.005]$} & \multirow{6}{*}{$[0.995, 1.005]$} \\
& BO (RBF)         & $0.992$ & $0.992$ & & \\
& BO (Matern $\nu=1.5$) & $0.991$ & $0.993$ & & \\
& BO (Matern $\nu=2.5$) & $0.994$ & $0.994$ & & \\
& MLE             & $1.029$ & $\mathbf{1.002}$ & & \\
& EM              & $1.134$ & $1.147$ & & \\
\hline
\end{tabular}
\label{tab:consistency_comparison_merged_full}
\end{table*}

\subsection{Second-order Model Experiments}

\begin{table*}[htbp]
\centering
\caption{PARAMETER ESTIMATION RESULTS FOR {\bf Scenario \textcircled{e}} (bold denotes the smallest RMSE across baselines) }
\renewcommand{\arraystretch}{1.3}
\begin{tabular}{|c|c|c|c|c|c|c|}
\hline
\textbf{Method} & \textbf{Parameter} & \textbf{Average Estimation} & \textbf{RMSE} & \textbf{NEES} & \textbf{NIS} & \textbf{Average log-likelihood}\\
\hline
\multirow{3}{*}{BO (EGP) } 
& $a_1$ & $4.98$ & $7.64e{-1}$ & \multirow{4}{*}{\centering 2.007} & \multirow{4}{*}{\centering 0.996} & \multirow{4}{*}{\centering -73.59} \\
& $a_0$ & $3.00$ & $4.32e{-1}$ &  &  &\\
& $Q$ & $2.01e{-2}$ & $2.69e{-3}$ &  &  &\\
& $R$ & $5.00e{-2}$ & $4.66e{-3}$ &  &  &\\
\hline
\multirow{3}{*}{BO (RBF)} 
& $a_1$ & $5.04$ & $1.27$ & \multirow{4}{*}{\centering 2.000} & \multirow{4}{*}{\centering 0.999} & \multirow{4}{*}{\centering -81.05}\\
& $a_0$ & $3.02$ & $5.92e{-1}$ &  &  &\\
& $Q$ & $2.09e{-2}$ & $5.42e{-3}$ &  & & \\
& $R$ & $4.98e{-2}$ & $3.85e{-3}$ &  &  & \\
\hline
\multirow{3}{*}{BO (Matern $\nu=1.5$)} 
& $a_1$ & $5.11$ & $1.25$ & \multirow{4}{*}{\centering 2.008} & \multirow{4}{*}{\centering 1.001} & \multirow{4}{*}{\centering -81.24}\\
& $a_0$ & $3.01$ & $6.39e{-1}$ &  &  &\\
& $Q$ & $2.07e{-2}$ & $4.67e{-3}$ &  & & \\
& $R$ & $4.98e{-2}$ & $3.82e{-3}$ &  & & \\
\hline
\multirow{3}{*}{BO (Matern $\nu=2.5$)} 
& $a_1$ & $5.12$ & $1.23$ & \multirow{4}{*}{\centering 2.004} & \multirow{4}{*}{\centering 1.000} & \multirow{4}{*}{\centering -81.18} \\
& $a_0$ & $3.00$ & $5.42e{-1}$ &  &  &\\
& $Q$ & $2.10e{-2}$ & $5.68e{-3}$ &  &  &\\
& $R$ & $4.98e{-2}$ & $3.83e{-3}$ &  &  &\\
\hline
\multirow{3}{*}{MLE} 
& $a_1$ & $5.27$ & $1.30$ & \multirow{4}{*}{\centering 1.909} & \multirow{4}{*}{\centering 0.988} & \multirow{4}{*}{\centering -81.47}\\
& $a_0$ & $3.08$ & $4.09e{-1}$ &  &  &\\
& $Q$ & $2.23e-2$ & $7.50e-3$ &  & & \\
& $R$ & $5.07e-2$ & $1.38e-3$ &  &  & \\
\hline
\multirow{3}{*}{EM} 
& $a_1$ & $5.73$ & $8.80e{-1}$ & \multirow{4}{*}{\centering 1.813} & \multirow{4}{*}{\centering 0.986} & \multirow{4}{*}{\centering -78.51}\\
& $a_0$ & $2.62$ & $4.43e{-1}$ &  &  &\\
& $Q$ & $2.50e{-2}$ & $6.37e{-3}$ &  & & \\
& $R$ & $4.95e{-2}$ & $2.44e{-3}$ &  &  &\\
\hline
\multicolumn{6}{l}{\small Units: $T$ ($\text{hr}$), $a$ ($\text{hr}^{-1}$), $Q$ and $R$ ($\text{deg}^2/\text{hr}^2$)}\\
\end{tabular}
\label{tab:scenario_e}
\end{table*}

\begin{table*}[htbp]
\centering
\caption{PARAMETER ESTIMATION RESULTS FOR {\bf Scenario \textcircled{f}} (bold denotes the smallest RMSE across baselines) }
\renewcommand{\arraystretch}{1.3}
\begin{tabular}{|c|c|c|c|c|c|c|}
\hline
\textbf{Method} & \textbf{Parameter} & \textbf{Average Estimation} & \textbf{RMSE} & \textbf{NEES} & \textbf{NIS} & \textbf{Average log-likelihood}\\
\hline
\multirow{3}{*}{BO (EGP) } 
& $a_1$ & $7.02$ & $5.64e{-1}$ & \multirow{4}{*}{\centering 1.999} & \multirow{4}{*}{\centering 0.997} & \multirow{4}{*}{\centering -152.89} \\
& $a_0$ & $1.99$ & $2.18e{-1}$ &  &  &\\
& $Q$ & $2.02e{-2}$ & $2.31e{-3}$ &  &  &\\
& $R$ & $6.03e{-2}$ & $4.43e{-3}$ &  &  &\\
\hline
\multirow{3}{*}{BO (RBF)} 
& $a_1$ & $7.25$ & $1.18$ & \multirow{4}{*}{\centering 1.959} & \multirow{4}{*}{\centering 0.996} & \multirow{4}{*}{\centering -159.23}\\
& $a_0$ & $2.10$ & $4.95e{-1}$ &  &  &\\
& $Q$ & $2.09e{-2}$ & $5.51e{-3}$ &  & & \\
& $R$ & $6.02e{-2}$ & $6.59e{-3}$ &  &  & \\
\hline
\multirow{3}{*}{BO (Matern $\nu=1.5$)} 
& $a_1$ & $7.07$ & $1.02$ & \multirow{4}{*}{\centering 1.991} & \multirow{4}{*}{\centering 1.001} & \multirow{4}{*}{\centering -158.75}\\
& $a_0$ & $2.04$ & $3.18e{-1}$ &  &  &\\
& $Q$ & $2.03e{-2}$ & $3.09e{-3}$ &  & & \\
& $R$ & $6.00e{-2}$ & $6.41e{-3}$ &  & & \\
\hline
\multirow{3}{*}{BO (Matern $\nu=2.5$)} 
& $a_1$ & $7.04$ & $9.80e{-1}$ & \multirow{4}{*}{\centering 1.989} & \multirow{4}{*}{\centering 1.000} & \multirow{4}{*}{\centering 158.81} \\
& $a_0$ & $2.03$ & $3.056e{-1}$ &  &  &\\
& $Q$ & $2.03e{-2}$ & $3.13e{-3}$ &  &  &\\
& $R$ & $6.00e{-2}$ & $6.44e{-3}$ &  &  &\\
\hline
\multirow{3}{*}{MLE} 
& $a_1$ & $5.27$ & $1.30$ & \multirow{4}{*}{\centering 1.909} & \multirow{4}{*}{\centering 0.988} & \multirow{4}{*}{\centering -181.06}\\
& $a_0$ & $3.08$ & $4.09e{-1}$ &  &  &\\
& $Q$ & $2.23e-2$ & $7.50e-3$ &  & & \\
& $R$ & $5.07e-2$ & $1.38e-3$ &  &  & \\
\hline
\multirow{3}{*}{EM} 
& $a_1$ & 7.66 & $8.98e{-1}$ & \multirow{4}{*}{\centering 1.842} & \multirow{4}{*}{\centering 0.992} & \multirow{4}{*}{\centering -151.47}\\
& $a_0$ & $2.51$ & $5.11e{-1}$ &  &  &\\
& $Q$ & $2.40e{-2}$ & $4.57e{-3}$ &  & & \\
& $R$ &  $5.95e{-2}$& $2.90e{-3}$ &  &  &\\
\hline
\multicolumn{6}{l}{\small Units: $T$ ($\text{hr}$), $a$ ($\text{hr}^{-1}$), $Q$ and $R$ ($\text{deg}^2/\text{hr}^2$)}\\
\end{tabular}
\label{tab:scenario_f}
\end{table*}

\begin{table*}[htbp]
\centering
\caption{CONSISTENCY TEST RESULTS FOR 10 HR FROM 100 MC TRIALS UNDER DIFFERENT SETTINGS (bold = within region)}
\renewcommand{\arraystretch}{1.2}
\begin{tabular}{|c|c|c|c|c|c|}
\hline
\textbf{Setting} & \textbf{Method} & \textbf{NEES} & \textbf{NIS} & \textbf{NEES Region(90\%)} & \textbf{NIS Region(90\%)} \\
\hline
\multirow{6}{*}{\textbf{\textcircled{e}}} & BO (EGP)  & $\mathbf{2.007}$ & $\mathbf{0.996}$ & \multirow{6}{*}{$[1.990, 2.010]$} & \multirow{6}{*}{$[0.993, 1.007]$} \\
& BO (RBF) & $\mathbf{2.000}$ & $\mathbf{0.999}$ & & \\
& BO (Matern $\nu=1.5$) & $\mathbf{2.008}$ & $\mathbf{1.001}$ & & \\
& BO (Matern $\nu=2.5$) & $\mathbf{2.004}$ & $\mathbf{1.000}$ & & \\
& MLE & $1.909$ & $0.988$ & & \\
& EM & $1.813$ & $0.986$ & & \\
\hline
\multirow{6}{*}{\textbf{\textcircled{f}}} & BO (EGP)  & $\mathbf{1.999}$ & $\mathbf{0.997}$ & \multirow{6}{*}{$[1.990, 2.010]$} & \multirow{6}{*}{$[0.993, 1.007]$} \\
& BO (RBF) & $1.959$ & $\mathbf{0.996}$ & & \\
& BO (Matern $\nu=1.5$) & $\mathbf{1.991}$ & $\mathbf{1.001}$ & & \\
& BO (Matern $\nu=2.5$) & $1.989$ & $\mathbf{1.000}$ & & \\
& MLE & $1.909$ & $0.988$ & & \\
& EM & $1.842$ & $0.992$ & & \\
\hline
\end{tabular}
\label{tab:consistency_comparison_merged_full_final}
\end{table*}
The second-order model experiments extend the analysis to a more complex parameter identification problem involving a second-order stochastic differential equation as described in Section~\ref{subsec:second_order_model}. Unlike the first-order case with three unknowns ($a$, $Q$, $R$), the second-order model requires estimating four parameters ($a_0$, $a_1$, $\tilde{Q}$, $R$), where $a_0$ and $a_1$ determine the system's characteristic polynomial as shown in Eq.~\eqref{eq:second_order_sde}. The state vector $\mathbf{x} = [x, \dot{x}]^T$ evolves according to the discrete-time model in Eq.~\eqref{eq:second_order_discrete}, with process noise covariance structure given by Eq.~\eqref{eq:Q_second_order_final}.

Tables~\ref{tab:scenario_e}--\ref{tab:scenario_f} present the estimation results across 100 Monte Carlo runs for Scenarios \textcircled{e}--\textcircled{f}. In Scenario \textcircled{e}, the coupled nature of parameters $a_0$ and $a_1$ in the state transition matrix poses significant challenges. BO (EGP)  achieves RMSEs of $4.32\times10^{-1}$ and $7.64\times10^{-1}$ for $a_0$ and $a_1$ respectively, maintaining balanced performance across all parameters. In contrast, MLE shows degraded performance for $a_1$ with RMSE of $1.30$, despite achieving slightly better $a_0$ estimation ($4.09\times10^{-1}$). The EM algorithm exhibits systematic bias, particularly evident in its $a_0$ estimate (2.62 versus true value 3.00). For Scenario \textcircled{f} with different stability characteristics ($a_0=7$, $a_1=2$), BO (EGP) demonstrates clear superiority, achieving the lowest RMSEs for both dynamic parameters—$2.18\times10^{-1}$ for $a_0$ and $5.64\times10^{-1}$ for $a_1$—while maintaining comparable accuracy for noise parameters $\tilde{Q}$ and $R$.

The filter consistency analysis for the second-order system reveals distinct statistical properties. With a two-dimensional state vector, the NEES metric computed via Eq.~\eqref{eq:NEES_vector} now follows a chi-squared distribution with two degrees of freedom per time step, yielding theoretical acceptance regions of $[1.990, 2.010]$ for the averaged NEES (90\% confidence). As shown in Table~\ref{tab:consistency_comparison_merged_full_final}, BO (EGP)  maintains excellent consistency with NEES values of 2.007 and 1.999 for Scenarios \textcircled{e} and \textcircled{f} respectively, closely matching the theoretical mean of 2.0. This indicates proper uncertainty quantification despite the increased state dimensionality. MLE and EM, however, show systematic underconfidence with NEES values below 1.95, suggesting overestimation of state uncertainties. The NIS values, still computed for scalar observations via Eq.~\eqref{eq:NIS_scalar}, remain near unity across all methods, confirming accurate measurement prediction.

The increased complexity of the second-order model—with its coupled dynamics, higher-dimensional state space, and additional parameters—amplifies the advantages of BO's global optimization strategy. The GP-based surrogate effectively captures the complex interactions between $a_0$ and $a_1$ that arise from the nilpotent approximation in Eq.~\eqref{eq:nilpotent_approx}, where the process noise covariance exhibits coupling between position and velocity states. This superior parameter identification translates to better log-likelihood values (BO (EGP)  achieves -73.59 versus MLE's -81.47 in Scenario \textcircled{e}) and robust filter consistency, demonstrating that the BO framework scales effectively to higher-dimensional parameter estimation problems while maintaining statistical rigor.

\section{Conclusions}\label{sec:conclusions}
This paper presents a Bayesian Optimization framework with ensemble Gaussian process kernels for parameter estimation in linear stochastic dynamical systems. By treating the log-likelihood function as a black box and employing multiple kernel functions with adaptive weighting, the proposed approach successfully identifies parameters in both first-order (Ornstein-Uhlenbeck) and higher-order stochastic differential equation models.

Extensive simulation results across six scenarios demonstrate that the ensemble BO approach consistently yields the lowest overall RMSEs compared to classical MLE and EM methods. For first-order OU models, the method achieves remarkable accuracy for the notoriously challenging inverse time constant. For second-order systems with coupled dynamics and four unknown parameters ($a_0$, $a_1$, $\tilde{Q}$, $R$), the ensemble BO effectively handles the increased complexity and parameter interactions arising from the nilpotent approximation structure. Notably, estimation errors fall below the Cramér-Rao Lower Bound—an apparent paradox arising from the implicit parameter prior induced by the GP ensemble. The adaptive weight mechanism successfully identifies the most suitable kernel for each scenario, with EGP dominating for complex, multi-modal likelihood surfaces while other kernels contribute in smoother parameter regimes.

The ensemble formulation represents a principled compromise between purely model-based approaches (requiring explicit likelihood gradients) and purely data-driven methods (ignoring system structure). By leveraging GP flexibility while incorporating multiple kernel hypotheses, we achieve automatic adaptation to varying system orders and parameter dimensionality without manual kernel tuning. The method maintains excellent filter consistency across both first- and second-order models, with NEES values closely matching theoretical expectations despite the different state dimensions.

However, the violation of the CRLB leaves the BO-based estimator without an explicit variance bound, critical for safety-critical applications. Future work will focus on establishing theoretical variance bounds for ensemble BO estimators, extending the framework to nonlinear stochastic dynamical systems, and exploring structured kernel dictionaries tailored to different model orders. Additionally, investigating the scalability to higher-dimensional systems and online weight adaptation strategies for real-time parameter tracking presents promising research directions.

\appendix

\subsection{EM Approach for High-Order OU Model Identification}

In this section, we present the EM approach to find the MLE of the OU model parameters $\bbtheta$ by maximizing the LLF~\eqref{eq:LLF}. The EM algorithm alternates between estimating the conditional expectation (E-step) and maximizing this expectation with respect to the model parameters (M-step).

\noindent\textbf{Complete data log-likelihood:}  
For a general $d$-dimensional state-space model,
\begin{align}
\mathbf{x}_n &= \mathbf{A}\,\mathbf{x}_{n-1} + \mathbf{v}_n, \quad \mathbf{v}_n \sim \mathcal{N}(\mathbf{0},\mathbf{Q}) \\
z_n &= \mathbf{H}\,\mathbf{x}_n + w_n, \quad w_n \sim \mathcal{N}(0,R)
\end{align}
the complete data log-likelihood is
\begin{align}
\log p(\mathbf{X},\mathbf{z};\bbtheta) 
&= \sum_{n=1}^N \Big[ \log p(\mathbf{x}_n|\mathbf{x}_{n-1};\bbtheta) + \log p(z_n|\mathbf{x}_n;\bbtheta) \Big] \nonumber\\
&= \sum_{n=1}^N \Big[  - \tfrac{1}{2}(\mathbf{x}_n - \mathbf{A}\mathbf{x}_{n-1})^\mathsf{T} \mathbf{Q}^{-1} (\mathbf{x}_n - \mathbf{A}\mathbf{x}_{n-1}) \nonumber\\
&\quad -\tfrac{1}{2}\log|2\pi \mathbf{Q}| -\tfrac{1}{2}\log(2\pi R) \nonumber\\
&\quad -\tfrac{1}{2}(z_n - \mathbf{H}\mathbf{x}_n)^\mathsf{T} R^{-1} (z_n - \mathbf{H}\mathbf{x}_n) \Big] 
\end{align}
\subsubsection{E-Step}
The E-step objective is
\begin{align}
U(\bbtheta; \bbtheta_i) := \mathbb{E}_{p(\mathbf{X}|\mathbf{z};\bbtheta_i)} \big[ \log p(\mathbf{X},\mathbf{z};\bbtheta) \big] \label{eq:U_high}
\end{align}
where the joint state posterior $p(\mathbf{X}|\mathbf{z};\bbtheta_i)$ is obtained from the Kalman smoother given the parameter estimate $\bbtheta_i$ from the previous iteration.

\noindent\textbf{Forward filtering:} Implemented as in Alg.~\ref{table:loss}, but with $\mathbf{x}_n \in \mathbb{R}^d$ and $\mathbf{P}_{n|n} \in \mathbb{R}^{d\times d}$.  

\noindent\textbf{Backward smoothing}  
\begin{subequations}
\begin{align}
\mathbf{J}_n &= \mathbf{P}_{n|n} \mathbf{A}^\mathsf{T} \mathbf{P}_{n+1|n}^{-1} \\
\hat{\mathbf{x}}_{n|N} &= \hat{\mathbf{x}}_{n|n} + \mathbf{J}_n \big( \hat{\mathbf{x}}_{n+1|N} - \hat{\mathbf{x}}_{n+1|n} \big) \\
\mathbf{P}_{n|N} &= \mathbf{P}_{n|n} + \mathbf{J}_n \big( \mathbf{P}_{n+1|N} - \mathbf{P}_{n+1|n} \big) \mathbf{J}_n^\mathsf{T}
\end{align}
\end{subequations}
We also compute the lag-one smoothed covariances
\[
\mathbf{P}_{n,n-1|N} = \mathbb{E}\big[(\mathbf{x}_n - \hat{\mathbf{x}}_{n|N})(\mathbf{x}_{n-1} - \hat{\mathbf{x}}_{n-1|N})^\mathsf{T} \big].
\]

With these quantities,~\eqref{eq:U_high} can be expressed as
\begin{align}
&U(\bbtheta; \bbtheta_i) = -\frac{N}{2}\log|2\pi \mathbf{Q}| - \frac{1}{2} \sum_{n=1}^N \mathrm{tr}\big[ \mathbf{Q}^{-1} \mathbf{E}_n \big] \nonumber\\
&\quad -\frac{N}{2}\log(2\pi R) - \frac{1}{2R} \sum_{n=1}^N \mathbb{E}\big[ (z_n - \mathbf{H}\mathbf{x}_n)^2 \big]
\end{align}
where
\[
\mathbf{E}_n = \mathbf{P}_{n|N} + \hat{\mathbf{x}}_{n|N} \hat{\mathbf{x}}_{n|N}^\mathsf{T}
- \mathbf{A}\mathbf{P}_{n-1|N} \mathbf{A}^\mathsf{T}
- \mathbf{A}\hat{\mathbf{x}}_{n-1|N}\hat{\mathbf{x}}_{n-1|N}^\mathsf{T}.
\]

\subsubsection{M-Step}
Maximizing $U(\bbtheta; \bbtheta_i)$ with respect to $\mathbf{A}$, $\mathbf{Q}$, and $R$ yields

\noindent\textbf{Update for $\mathbf{A}$:}
\begin{align}
\mathbf{A}_{i+1} &= \left( \sum_{n=1}^N \big[ \mathbf{P}_{n,n-1|N} + \hat{\mathbf{x}}_{n|N} \hat{\mathbf{x}}_{n-1|N}^\mathsf{T} \big] \right) \nonumber\\
&\quad \times \left( \sum_{n=1}^N \big[ \mathbf{P}_{n-1|N} + \hat{\mathbf{x}}_{n-1|N} \hat{\mathbf{x}}_{n-1|N}^\mathsf{T} \big] \right)^{-1}
\end{align}

\noindent\textbf{Update for $\mathbf{Q}$}
\begin{align}
\mathbf{Q}_{i+1} &= \frac{1}{N} \sum_{n=1}^N \Big[ \mathbf{P}_{n|N} + \hat{\mathbf{x}}_{n|N}\hat{\mathbf{x}}_{n|N}^\mathsf{T} \nonumber\\
&\quad - \mathbf{A}_{i+1} \big( \mathbf{P}_{n,n-1|N} + \hat{\mathbf{x}}_{n|N}\hat{\mathbf{x}}_{n-1|N}^\mathsf{T} \big)^\mathsf{T} \Big]
\end{align}

\noindent\textbf{Update for $R$}
\begin{align}
R_{i+1} &= \frac{1}{N} \sum_{n=1}^N \Big[ (z_n - \mathbf{H}\hat{\mathbf{x}}_{n|N})^2 + \mathbf{H} \mathbf{P}_{n|N} \mathbf{H}^\mathsf{T} \Big]
\end{align}

\subsubsection{Learning-Rate Updates}
We apply learning rate updates for stability
\begin{subequations}
\begin{align}
\mathbf{A}_{i+1} &\leftarrow (1-\alpha)\mathbf{A}_{i} + \alpha\,\mathbf{A}_{i+1} \\
\mathbf{Q}_{i+1} &\leftarrow (1-\alpha)\mathbf{Q}_{i} + \alpha\,\mathbf{Q}_{i+1} \\
R_{i+1} &\leftarrow (1-\alpha)R_{i} + \alpha\,R_{i+1}
\end{align}
\end{subequations}
where $\alpha\in[0,1]$ controls the trade-off between stability and adaptation speed.

\begin{algorithm}[htbp]
\caption{EM Algorithm for High-Order OU / Linear Gaussian SSM}
\label{alg:em-high}
\begin{algorithmic}[1]
\STATE \textbf{Input:} Observations ${\bf z}_N$,
Initial parameters $\bbtheta_0 = [\mathbf{A}^{(0)}, \mathbf{Q}^{(0)}, R^{(0)}]$,
Learning rate $\alpha$,
Tolerance $\varepsilon$,
Max iterations $I_{\max}$.
\FOR{$i = 0$ \TO $I_{\max}-1$}
\STATE \textbf{E-step:} Run Kalman smoother to obtain $\{\hat{\mathbf{x}}_{n|N}, \mathbf{P}_{n|N}, \mathbf{P}_{n,n-1|N}\}$
\STATE \textbf{M-step:} Update $\mathbf{A}$, $\mathbf{Q}$, and $R$ using the above formulas
\STATE \textbf{Learning-rate update:} Apply learning rate to $\mathbf{A}$, $\mathbf{Q}$, and $R$
\STATE \textbf{Convergence check:}
\IF{$\|\bbtheta_{i+1} - \bbtheta_i\| < \varepsilon$}
\STATE \textbf{break}
\ENDIF
\ENDFOR
\STATE \textbf{Output:} $\bbtheta^* \leftarrow \bbtheta_{i+1}$
\end{algorithmic}
\end{algorithm}

\subsection{Statistical Distribution of Consistency Metrics}

This appendix derives the theoretical distributions for the averaged NEES and NIS statistics used in Section~\ref{subsec:metrics} for filter consistency validation.

\subsubsection{Distribution of Normalized Estimation Error Squared (NEES)}

For a $d$-dimensional state vector, the NEES at time $n$ for the $j$th Monte Carlo run is defined in Eq.~\eqref{eq:NEES_vector} as
\begin{align}
\epsilon_n^{(j)} = (\mathbf{x}_n^{(j)} - \hat{\mathbf{x}}_{n|n}^{(j)})^\top (\mathbf{P}_{n|n}^{(j)})^{-1} (\mathbf{x}_n^{(j)} - \hat{\mathbf{x}}_{n|n}^{(j)})
\end{align}

Under the assumption that the filter is consistent (i.e., the estimated parameters equal the true parameters), the estimation error follows
\begin{align}
\mathbf{x}_n^{(j)} - \hat{\mathbf{x}}_{n|n}^{(j)} \sim \mathcal{N}(\mathbf{0}, \mathbf{P}_{n|n}^{(j)})
\end{align}

The quadratic form $\epsilon_n^{(j)}$ then follows a chi-squared distribution with $d$ degrees of freedom
\begin{align}
\epsilon_n^{(j)} \sim \chi^2_d
\end{align}

For the scalar case ($d=1$) in first-order models, this reduces to $\epsilon_n^{(j)} \sim \chi^2_1$, while for second-order models with $d=2$, we have $\epsilon_n^{(j)} \sim \chi^2_2$.

The averaged NEES over all Monte Carlo runs and time steps
\begin{align}
\bar{\epsilon} = \frac{1}{N_{\rm MC}N} \sum_{j=1}^{N_{\rm MC}}\sum_{n=1}^N \epsilon_n^{(j)}
\end{align}

Since the $\epsilon_n^{(j)}$ are independent and identically distributed, their sum follows
\begin{align}
\sum_{j=1}^{N_{\rm MC}}\sum_{n=1}^N \epsilon_n^{(j)} \sim \chi^2_{dN_{\rm MC}N}
\end{align}

Therefore, the normalized average follows
\begin{align}
\bar{\epsilon} \sim \frac{\chi^2_{dN_{\rm MC}N}}{N_{\rm MC}N}
\end{align}

with expected value $\mathbb{E}[\bar{\epsilon}] = d$ and variance $\text{Var}[\bar{\epsilon}] = 2d/(N_{\rm MC}N)$.

\subsubsection{Distribution of Normalized Innovation Squared (NIS)}

Similarly, for the NIS with $m$-dimensional observations, defined in Eq.~\eqref{eq:NIS_vector}
\begin{align}
\nu_n^{(j)} = (\mathbf{z}_n^{(j)} - \hat{\mathbf{z}}_{n|n-1}^{(j)})^\top (\mathbf{S}_{n|n-1}^{(j)})^{-1} (\mathbf{z}_n^{(j)} - \hat{\mathbf{z}}_{n|n-1}^{(j)})
\end{align}

Under filter consistency, the innovation follows
\begin{align}
\mathbf{z}_n^{(j)} - \hat{\mathbf{z}}_{n|n-1}^{(j)} \sim \mathcal{N}(\mathbf{0}, \mathbf{S}_{n|n-1}^{(j)})
\end{align}

Thus $\nu_n^{(j)} \sim \chi^2_m$. For scalar observations ($m=1$) used in our experiments, $\nu_n^{(j)} \sim \chi^2_1$.

The averaged NIS
\begin{align}
\bar{\nu} \sim \frac{\chi^2_{mN_{\rm MC}N}}{N_{\rm MC}N}
\end{align}

with $\mathbb{E}[\bar{\nu}] = m$ and $\text{Var}[\bar{\nu}] = 2m/(N_{\rm MC}N)$.

\subsubsection{Acceptance Regions}

For hypothesis testing at significance level $\alpha$, the $(1-\alpha)$ acceptance region is constructed as
\begin{align}
\left[\frac{\chi^2_{dN_{\rm MC}N,\alpha/2}}{N_{\rm MC}N}, \frac{\chi^2_{dN_{\rm MC}N,1-\alpha/2}}{N_{\rm MC}N}\right]
\end{align}

For our experiments with $N_{\rm MC}=100$ and varying $N$
\begin{itemize}
    \item First-order models ($d=1$): $\bar{\epsilon}$ should be near 1
    \item Second-order models ($d=2$): $\bar{\epsilon}$ should be near 2
    \item All models with scalar observations ($m=1$): $\bar{\nu}$ should be near 1
\end{itemize}

These theoretical distributions provide the basis for the consistency validation presented in Tables~\ref{tab:consistency_comparison_merged_full} and \ref{tab:consistency_comparison_merged_full_final}.
\section*{ACKNOWLEDGMENT}

The preferred spelling of the word ``acknowledgment'' in American English is without an ``e'' after the ``g.'' Use the singular heading even if you have many acknowledgments. Avoid expressions such as ``One of us (S.B.A.) would like to thank ... .'' Instead, write ``F. A. Author thanks ... .'' In most cases, sponsor and financial support acknowledgments are placed in the unnumbered footnote on the first page, not here.

\bibliographystyle{IEEEtran}  
\bibliography{BO_fusion}  

\begin{IEEEbiography}[{\includegraphics[width=1in,height=1.25in]{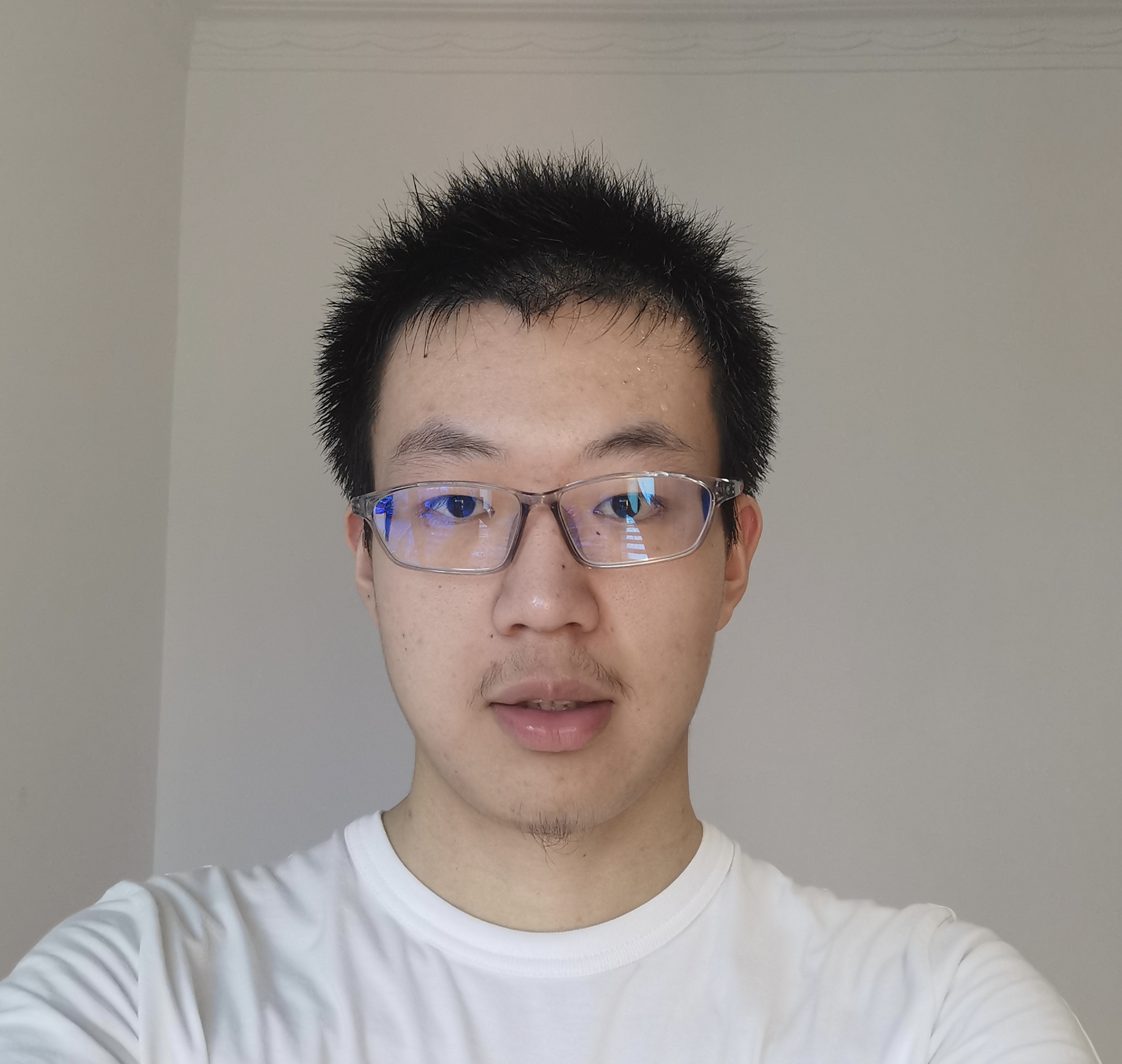}}]{Jinwen Xu}(Student Member, IEEE) received the B.S. degree from Nankai University, Tianjin, China, in 2023, and the M.S. degree from the University of Wisconsin-Madison, Madison, WI, USA, in 2024. He is currently pursuing the Ph.D. degree with the University of Georgia, Athens, GA, USA. His research interests include machine learning, data science, Gaussian processes, Bayesian optimization, and uncertainty quantification through hypothesis testing. His current work focuses on developing advanced statistical methods for model validation and parameter estimation in stochastic systems.
\end{IEEEbiography}%

\begin{IEEEbiography}[{\includegraphics[width=1in,height=1.25in]{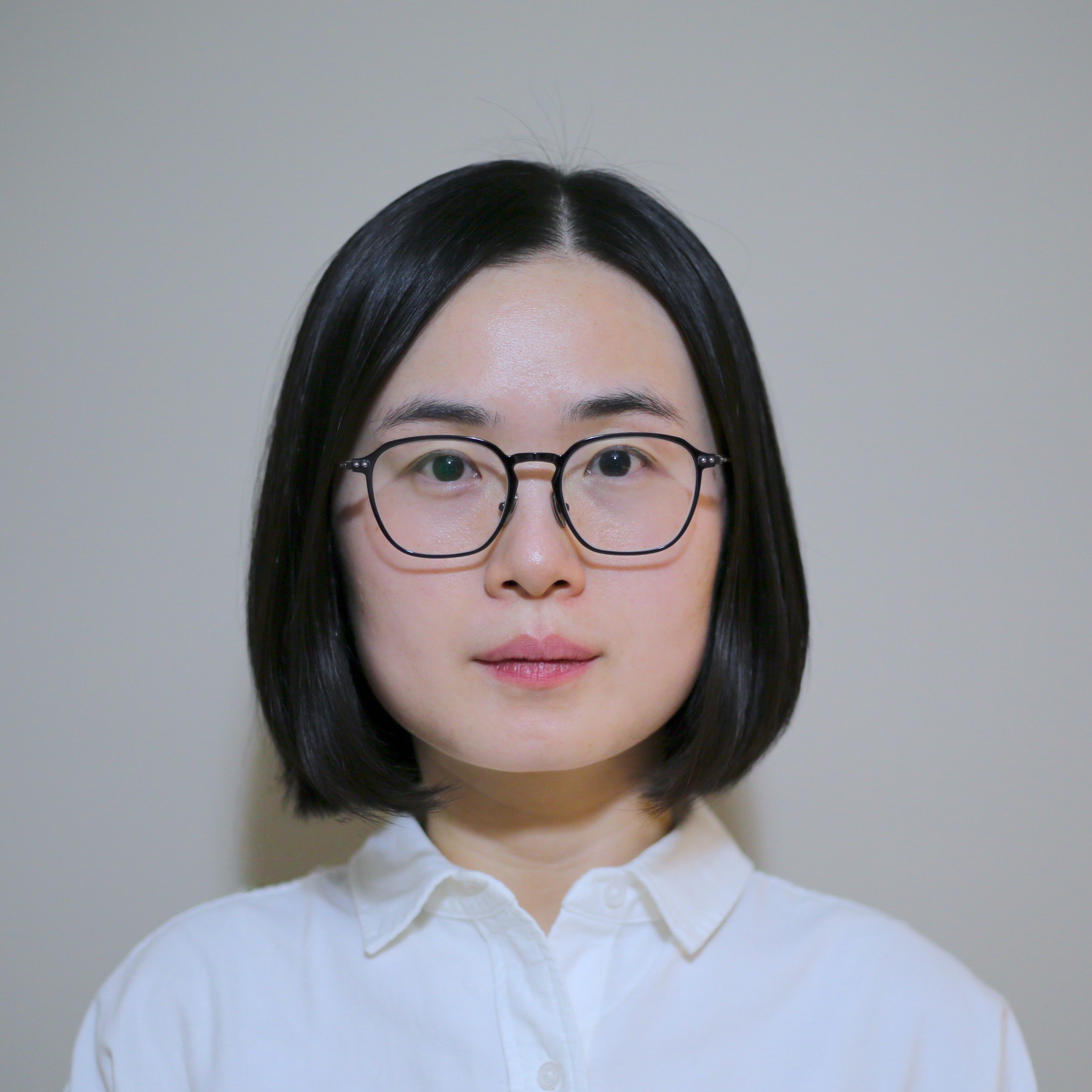}}]{Qin Lu}(Member, IEEE) received the B.S. degree from the University of Electronic Science and Technology of China in 2013 and the Ph.D. degree from the University of Connecticut (UConn) in 2018. Following the post-doctoral training at the University of Minnesota, she joined the School of Electrical and Computer Engineering at the University of Georgia as an Assistant Professor in 2023. Her research interests are in the areas of signal processing, machine learning, data science, and communications, with special focus on Gaussian processes, Bayesian optimization, spatio-temporal inference over graphs, and data association for multi-object tracking. She was awarded the Summer Fellowship and Doctoral Dissertation Fellowship from UConn. She was also a recipient of the Women of Innovation Award by Connecticut Technology Council in 2018, the NSF CAREER Award in 2024, and Best Student Paper Award in IEEE Sensor Array and Multichannel Workshop in 2024, and the UConn Engineering GOLD Rising Star Alumni Award in 2025.
\end{IEEEbiography}

\begin{IEEEbiography}[{\includegraphics[width=1in,height=1.25in]{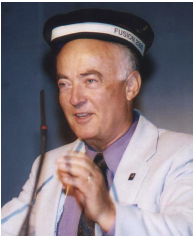}}]{Yaakov Bar-Shalom} (Life Fellow, IEEE)born in 1941. He received the B.S. and M.S. degrees from the Technion—Israel Institute of Technology, Haifa, Israel, 1963 and 1967, respectively, and the Ph.D. degree from Princeton University, Princeton, NJ, USA, in 1970, all in electrical engineering.
He is currently a Board of Trustees Distinguished Professor with the Department of Electronics and Communication Engineering. and a M. E. Klewin Professor with the University of
Connecticut (UConn), Mansfield, CT, USA. His current research interests are in estimation theory, target tracking, and data fusion. He has authored or coauthored more than 650 papers and book chapters in these areas and in stochastic adaptive control and eight books, including Estimation with Applications to Tracking and Navigation (Wiley 2001) and Tracking and Data Fusion (2011). He is currently an Associate Editor for IEEE TRANSACTIONS ON AUTOMATIC CONTROL and Automatica, General Chairman of 1985 ACC, FUSION 2000, and was ISIF President (2000, 2002) and VP Publications (2004–2013). He graduated 42 Ph.D.s at UConn and served as Co-major advisor for 6 Ph.D. degrees awarded elsewhere. He is co-recipient of the M. Barry Carlton Award for the best paper in IEEE TRANSACTIONS ON AEROSPACE AND ELECTRONIC SYSTEMS (1995, 2000), the 2022 IEEE Aerospace and Electronic Systems Society Pioneer Award and recipient of the 2008 IEEE Dennis J. Picard Medal for Radar Technologies and Applications and the 2012 Connecticut Medal of Technology. He has been listed by academic.research.microsoft as \#1 in Aerospace Engineering based on the citations of his work and is the recipient of the 2015 ISIF Award for a Lifetime of Excellence in Information Fusion, renamed in 2016 as ``ISIF Yaakov Bar-Shalom Award for Lifetime of Excellence in Information Fusion." He is also recipient (with H.A.P. Blom) of the 2022 IEEE AESS Pioneer Award for the IMM Estimator.
\end{IEEEbiography}

\end{document}